\documentclass[preprint,5p,compress]{elsarticle}

\usepackage{lscape,caption} 
\usepackage{amssymb}
\usepackage{amsmath}
\usepackage{mathtools, cuted}
\usepackage{graphicx}
\usepackage{url}
\usepackage{algorithm}
\usepackage{algpseudocode}
\usepackage{color}
\usepackage{booktabs}
\usepackage{stmaryrd}
\usepackage{epstopdf}
\usepackage{pdflscape}
\usepackage{stmaryrd}
\usepackage{color}
\usepackage[table]{xcolor}
\usepackage{array}
\usepackage{tikz}
\usepackage{caption}
\usepackage{subcaption}
\usepackage{dblfloatfix}
\usepackage{flushend}

\newcolumntype{V}{!{\vrule width 2pt}}

\journal{Information Fusion}

\begin{document}

\begin{frontmatter}

\title{A practical tutorial on autoencoders for nonlinear feature fusion: \\Taxonomy, models, software and guidelines}

\author[UGR]{David Charte\corref{cor1}}
\ead{fdavidcl@correo.ugr.es}

\author[UJA]{Francisco Charte}
\ead{fcharte@ujaen.es}

\author[UGR]{Salvador Garc\'ia}
\ead{salvagl@decsai.ugr.es}

\author[UJA]{Mar\'ia J. del Jesus}
\ead{mjjesus@ujaen.es}

\author[UGR,IND]{Francisco Herrera}
\ead{herrera@decsai.ugr.es}

\cortext[cor1]{Corresponding author. Tel.: +34~953~213~016\\%
  Manuscript accepted at Information Fusion: \url{https://doi.org/10.1016/j.inffus.2017.12.007}.\\%
  \textcopyright~2018. This manuscript version is made available under the CC-BY-NC-ND 4.0 license \url{http://creativecommons.org/licenses/by-nc-nd/4.0/}}
\address[UGR]{Department of Computer Science and A.I., University of Granada, 18071 Granada, Spain}
\address[UJA]{Department of Computer Science, University of Ja\'en, 23071 Ja\'en, Spain}
\address[IND]{Faculty of Computing and Information Technology, King Abdulaziz University, 21589, Jeddah, Saudi Arabia}

\begin{abstract}
Many of the existing machine learning algorithms, both supervised and unsupervised, depend on the quality of the input characteristics to generate a good model. The amount of these variables is also important, since performance tends to decline as the input dimensionality increases, hence the interest in using feature fusion techniques, able to produce feature sets that are more compact and higher level. A plethora of procedures to fuse original variables for producing new ones has been developed in the past decades. The most basic ones use linear combinations of the original variables, such as PCA (\textit{Principal Component Analysis}) and LDA (\textit{Linear Discriminant Analysis}), while others find manifold embeddings of lower dimensionality based on non-linear combinations, such as Isomap or LLE (\textit{Linear Locally Embedding}) techniques. 

More recently, autoencoders (AEs) have emerged as an alternative to manifold learning for conducting nonlinear feature fusion. Dozens of AE models have been proposed lately, each with its own specific traits. Although many of them can be used to generate reduced feature sets through the fusion of the original ones, there also AEs designed with other applications in mind.

The goal of this paper is to provide the reader with a broad view of what an AE is, how they are used for feature fusion, a taxonomy gathering a broad range of models, and how they relate to other classical techniques. In addition, a set of didactic guidelines on how to choose the proper AE for a given task is supplied, together with a discussion of the software tools available. Finally, two case studies illustrate the usage of AEs with datasets of handwritten digits and breast cancer.
\end{abstract}


\end{frontmatter}

\section{Introduction}
The development of the first machine learning techniques dates back to the middle of the 20th century, supported mainly by previously established statistical methods. By then, early research on how to emulate the functioning of the human brain through a machine was underway. McCulloch and Pitts cell \cite{McCullochPitts} was proposed back in 1943, and the Hebb rule \cite{HebbRule} that the Perceptron \cite{Perceptron} is founded on was stated in 1949. Therefore, it is not surprising that artificial neural networks (ANNs), especially since the backpropagation algorithm was rediscovered in 1986 by Rumelhart, Hinton and Willians \cite{Backpropagation}, have become one of the essential models. 

ANNs have been applied to several machine learning tasks, mostly following a supervised approach. As was mathematically demonstrated \cite{ANNsUniversalApproximator} in 1989, a multilayer feedforward ANN (MLP) is an universal approximator, hence their usefulness in classification and regression problems. However, a proper algorithm able to train an MLP with several hidden layers was not available, due to the vanishing gradient \cite{VanishingGradient} problem. The gradient descent algorithm, firstly used for convolutional neural networks \cite{LeCunBackpropagation} and later for unsupervised learning \cite{HintonDBN}, was one of the foundations of modern deep learning \cite{DeepLearning} methods.

Under the umbrella of deep learning, multiple techniques have emerged and evolved. These include DBNs (\textit{Deep Belief Networks}) \cite{DBNs}, CNNs (\textit{Convolutional Neural Networks}) \cite{CNNs}, RNNs (\textit{Recurrent Neural Networks}) \cite{RNN} as well as LSTMs (\textit{Long Short-Term Memory}) \cite{LSTMs} or AEs (\textit{autoencoders}).

The most common architecture in unsupervised deep learning is that of the \textit{encoder-decoder} \cite{EnergyUnsupervised}. Some techniques lack the encoder or the decoder and have to compute costly optimization algorithms to find a code or sampling methods to reach a reconstruction, respectively. Unlike those, AEs capture both parts in their structure, with the aim that training them becomes easier and faster. In general terms, AEs are ANNs which produce codifications for input data and are trained so that their decodifications resemble the inputs as closely as possible.

AEs were firstly introduced \cite{AEs} as a way of conducting pretraining in ANNs. Although mainly developed inside the context of deep learning, not all AE models are necessarily ANNs with multiple hidden layers. As explained below, an AE can be a deep ANN, i.e. in the stacked AEs configuration, or it can be a shallow ANN with a single hidden layer. See Section~\ref{Sec.Essentials} for a more detailed introduction to AEs.

While many machine learning algorithms are able to work with raw input features, it is also true that, for the most part, their behavior is degraded as the number of variables grows. This is mainly due to the problem known as the \textit{curse of dimensionality} \cite{Richard1957}, as well as the justification for a field of study called feature engineering. Engineering of features started as a manual process, relying in an expert able to decide by observation which variables were better for the task at hand. Notwithstanding, automated feature selection \cite{FeatureSelection} methods were soon available.

Feature selection is only one of the approaches to reduce input space dimensionality. Selecting the best subset of input variables is an NP-hard combinatorial problem. Moreover, feature selection techniques usually evaluate each variable independently, but it is known that variables that separately do not provide useful information may do so when they are used together. For this reason other alternatives, primarily feature construction or extraction \cite{FeatureExtraction}, emerged. In addition to these two denominations, feature selection and feature extraction, when dealing with dimensionality reduction it is also frequent to use other terms. The most common are as follows:

	\paragraph{Feature engineering \cite{Domingos2012AFU}}
	This is probably the broadest term, encompassing most of the others. Feature engineering can be carried out by manual or automated means, and be based on the selection of original characteristics or the construction of new ones through transformations. 

	\paragraph{Feature learning \cite{bengio_representation_2013}}
	It is the denomination used when the process to select among the existing features or construct new ones is automated. Thus, we can perform both feature selection and feature extraction through algorithms such as the ones mentioned below. Despite the use of automatic methods, sometimes an expert is needed to decide which algorithm is the most appropriate depending on data traits, to evaluate the optimum amount of variables to extract, etc.
	
    \paragraph{Representation learning \cite{bengio_representation_2013}}
    Although this term is sometimes interchangeably used with the previous one, it is mostly used to refer to the use of ANNs to fully automate the feature generation process. Applying ANNs to learn distributed representations of concepts was proposed by Hinton in \cite{DistributedRepresentations}. Today, learning representations is mainly linked to processing natural language, images and other signals with specific kinds of ANNs, such as CNNs \cite{CNNs}.
  
    \paragraph{Feature selection \cite{DataPreprocessing}} 
    Picking the most informative subset of variables started as a manual process usually in charge of domain experts. It can be considered a special case of feature weighting, as discussed in \cite{FeatureWeighting}. Although in certain fields the expert is still an important factor, nowadays the selection of variables is usually carried out using computer algorithms. These can operate in supervised or unsupervised manner. The former approach usually relies on correlation or mutual information between input and output variables \cite{CorrelationFS,MutualInformationDS}, while the latter tends to avoid redundancy among features \cite{UnsupervisedFS}. Feature selection is overall an essential strategy in the data preprocessing \cite{PreprocTutorial, DataPreprocessing} phase.
    
    \paragraph{Feature extraction \cite{FeatureExtractionIntro}} 
    The goal of this technique is to find a better data representation for the machine learning algorithm intended to use, since the original representation might not be the best one. It can be faced both manually, in which case the \textit{feature construction} term is of common use, and automatically. Some elemental techniques such as normalization, discretization or scaling of variables, as well as basic transformations applied to certain data types\footnote{e.g. Take the original field containing a date and divide it into three new variables, year, month and day.}, are also considered within this field. New features can be extracted by finding linear combinations of the original ones, as in PCA (\textit{Principal Component Analysis}) \cite{PCA,PCAHotelling} or LDA (\textit{Linear Discriminant Analysis}) \cite{LDA}, as well as nonlinear combinations, like Kernel PCA \cite{KernelPCA}  or Isomap \cite{Isomap}. The latter ones are usually known as \textit{manifold learning} \cite{ManifoldLearning} algorithms, and fall in the scope of nonlinear dimensionality reduction techniques \cite{NonlinearDimRec}. Feature extraction methods can also be categorized as supervised (e.g. LDA) or non-supervised (e.g. PCA).
    
    \paragraph{Feature fusion \cite{FeatureFusion}}
    This more recent term has emerged with the growth of multimedia data processing by machine learning algorithms, especially images, text and sound. As stated in \cite{FeatureFusion}, feature fusion methods aim to combine variables to remove redundant and irrelevant information. Manifold learning algorithms, and especially those based on ANNs, fall into this category. \\

    Among the existing AE models there are several that are useful to perform feature fusion. This is the aim of the most basic one, which can be extended via several regularizations and adaptations to different kinds of data. These options will be explored through the present work, whose aim is to provide the reader with a didactic review on the inner workings of these distinct AE models and the ways they can be used to learn new representations of data.

    The following are the main contributions of this paper:
    \begin{itemize}
	    \item A proposal of a global taxonomy of AEs dedicated to feature fusion.
	
	    \item Descriptions of these AE models including the necessary mathematical formulation and explanations.
	
	    \item A theoretical comparison between AEs and other popular feature fusion techniques.
	    
	    \item A comprehensive review of other AE models as well as their applications.
	      
	    \item A set of guidelines on how to design an AE, and several examples on how an AE may behave when its architecture and parameters are altered.

            \item A summary of the available software for creating deep learning models and specifically AEs.
    \end{itemize}

    Additionally, we provide a case study with the well known dataset MNIST \cite{MNIST}, which gives the reader some intuitions on the results provided by an AE with different architectures and parameters. The scrips to reproduce these experiments are provided in a repository, and their use will be further described in Section \ref{Sec.HowToChoose}.

    The rest of this paper is structured as follows. The foundations and essential aspects of AEs are introduced in Section \ref{Sec.Essentials}, including the proposal of a global taxonomy. Section \ref{Sect.AEforFF} is devoted to thoroughly describing the AE models able to operate as feature fusion mechanisms 
    and several models which have further applications. The relationship between these AE models and other feature fusion methods is portrayed in Section \ref{Sec.Relationship}, while applications of different kinds of AEs are described in Section \ref{Sec.Applications}. Section \ref{Sec.HowToChoose} provides a set of guidelines on how to design an AE for the task at hand, followed by the software pieces where it can be implemented, as well as the case study with MNIST data. Concluding remarks can be found in Section \ref{Conclusions}. Lastly, an Appendix briefly describes the datasets used through the present work.

\section{Autoencoder essentials}\label{Sec.Essentials}

AEs are ANNs\footnote{Strictly speaking not all AEs are ANNs, but here our interest is in those since they are the most common ones.} with a symmetric structure, where the middle layer represents an \textit{encoding} of the input data. AEs are trained to reconstruct their input onto the output layer, while verifying certain restrictions which prevent them from simply copying the data along the network. Although the term \textit{autoencoder} is the most popular nowadays, they were also known as autoassociative neural networks \cite{AutoassociativeNN}, diabolo networks \cite{DiaboloNN} and replicator neural networks \cite{ReplicatorNN}.

In this section the foundations of AEs are introduced, describing their basic architecture as ANNs as well as the activation functions regularly applied in their layers. Next, AEs are grouped into four types according to their architecture. This is followed by our proposed taxonomy for AEs, which takes into account the properties these induce in codifications. Lastly, a summary of their habitual applications is provided.

\subsection{General structure}
\begin{figure}[ht!]
\begin{center}
\begin{tikzpicture}[scale=0.15]
\tikzstyle{every node}+=[inner sep=0pt]
\draw [black] (21.3,-26.2) circle (3);
\draw (21.3,-26.2) node {$x$};
\draw [black] (31,-26.2) circle (3);
\draw (31,-26.2) node {$y$};
\draw [black] (40.5,-26.2) circle (3);
\draw (40.5,-26.2) node {$r$};
\draw [black] (24.3,-26.2) -- (28,-26.2);
\fill [black] (28,-26.2) -- (27.2,-25.7) -- (27.2,-26.7);
\draw (26.15,-25.7) node [above] {$f$};
\draw [black] (34,-26.2) -- (37.5,-26.2);
\fill [black] (37.5,-26.2) -- (36.7,-25.7) -- (36.7,-26.7);
\draw (35.75,-25.7) node [above] {$g$};
\end{tikzpicture}
\end{center}
  \caption{General autoencoder structure}
  \label{fig:structure}
\end{figure}
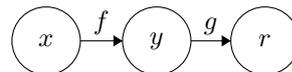

The basic structure of an AE, as shown in Fig.~\ref{fig:structure}, includes an input $x$ which is mapped onto the encoding $y$ via an encoder, represented as function $f$. This encoding is in turn mapped to the reconstruction $r$ by means of a decoder, represented as function $g$.

\begin{figure}[ht!]
  \centering
  \resizebox {0.85\columnwidth} {!} {
        \begin{tikzpicture}[scale=0.12]
\tikzstyle{every node}+=[inner sep=0pt]
\draw [black] (15.7,-15.6) circle (3);
\draw [black] (15.7,-23.8) circle (3);
\draw [black] (15.7,-31.9) circle (3);
\draw [black] (15.7,-39.6) circle (3);
\draw [black] (15.7,-47.6) circle (3);
\draw [black] (29.3,-23.8) circle (3);
\draw [black] (29.3,-31.9) circle (3);
\draw [black] (29.3,-39.6) circle (3);
\draw [black] (41.8,-27) circle (3);
\draw [black] (41.8,-36.5) circle (3);
\draw (23.5,-13.7) node {$W^{(1)}$};
\draw (35.5,-17.4) node {$W^{(2)}$};
\draw [black] (53.7,-23.8) circle (3);
\draw [black] (53.7,-31.9) circle (3);
\draw [black] (53.7,-39.6) circle (3);
\draw [black] (65.4,-15.6) circle (3);
\draw [black] (65.4,-23.8) circle (3);
\draw [black] (65.4,-31.9) circle (3);
\draw [black] (65.4,-39.6) circle (3);
\draw [black] (65.4,-47.6) circle (3);
\draw (47.7,-17.4) node {$W^{(3)}$};
\draw (58.3,-13.7) node {$W^{(4)}$};
\draw [black] (18.27,-17.15) -- (26.73,-22.25);
\fill [black] (26.73,-22.25) -- (26.3,-21.41) -- (25.79,-22.27);
\draw [black] (17.62,-17.9) -- (27.38,-29.6);
\fill [black] (27.38,-29.6) -- (27.25,-28.66) -- (26.48,-29.3);
\draw [black] (17.18,-18.21) -- (27.82,-36.99);
\fill [black] (27.82,-36.99) -- (27.86,-36.05) -- (26.99,-36.54);
\draw [black] (18.7,-23.8) -- (26.3,-23.8);
\fill [black] (26.3,-23.8) -- (25.5,-23.3) -- (25.5,-24.3);
\draw [black] (18.28,-25.34) -- (26.72,-30.36);
\fill [black] (26.72,-30.36) -- (26.29,-29.53) -- (25.78,-30.39);
\draw [black] (17.66,-26.07) -- (27.34,-37.33);
\fill [black] (27.34,-37.33) -- (27.2,-36.39) -- (26.44,-37.05);
\draw [black] (18.28,-30.36) -- (26.72,-25.34);
\fill [black] (26.72,-25.34) -- (25.78,-25.31) -- (26.29,-26.17);
\draw [black] (18.7,-31.9) -- (26.3,-31.9);
\fill [black] (26.3,-31.9) -- (25.5,-31.4) -- (25.5,-32.4);
\draw [black] (18.31,-33.38) -- (26.69,-38.12);
\fill [black] (26.69,-38.12) -- (26.24,-37.29) -- (25.75,-38.16);
\draw [black] (17.66,-37.33) -- (27.34,-26.07);
\fill [black] (27.34,-26.07) -- (26.44,-26.35) -- (27.2,-27.01);
\draw [black] (18.31,-38.12) -- (26.69,-33.38);
\fill [black] (26.69,-33.38) -- (25.75,-33.34) -- (26.24,-34.21);
\draw [black] (18.7,-39.6) -- (26.3,-39.6);
\fill [black] (26.3,-39.6) -- (25.5,-39.1) -- (25.5,-40.1);
\draw [black] (17.19,-45) -- (27.81,-26.4);
\fill [black] (27.81,-26.4) -- (26.98,-26.85) -- (27.85,-27.35);
\draw [black] (17.66,-45.33) -- (27.34,-34.17);
\fill [black] (27.34,-34.17) -- (26.43,-34.44) -- (27.19,-35.1);
\draw [black] (18.29,-46.08) -- (26.71,-41.12);
\fill [black] (26.71,-41.12) -- (25.77,-41.1) -- (26.28,-41.96);
\draw [black] (32.21,-24.54) -- (38.89,-26.26);
\fill [black] (38.89,-26.26) -- (38.24,-25.57) -- (37.99,-26.54);
\draw [black] (32.09,-30.81) -- (39.01,-28.09);
\fill [black] (39.01,-28.09) -- (38.08,-27.92) -- (38.44,-28.85);
\draw [black] (31.41,-37.47) -- (39.69,-29.13);
\fill [black] (39.69,-29.13) -- (38.77,-29.35) -- (39.48,-30.05);
\draw [black] (31.4,-25.94) -- (39.7,-34.36);
\fill [black] (39.7,-34.36) -- (39.49,-33.44) -- (38.78,-34.14);
\draw [black] (32.12,-32.94) -- (38.98,-35.46);
\fill [black] (38.98,-35.46) -- (38.41,-34.72) -- (38.06,-35.66);
\draw [black] (32.21,-38.88) -- (38.89,-37.22);
\fill [black] (38.89,-37.22) -- (37.99,-36.93) -- (38.23,-37.9);
\draw [black] (44.7,-37.26) -- (50.8,-38.84);
\fill [black] (50.8,-38.84) -- (50.15,-38.16) -- (49.9,-39.13);
\draw [black] (44.6,-35.42) -- (50.9,-32.98);
\fill [black] (50.9,-32.98) -- (49.98,-32.8) -- (50.34,-33.74);
\draw [black] (43.85,-34.31) -- (51.65,-25.99);
\fill [black] (51.65,-25.99) -- (50.74,-26.23) -- (51.47,-26.91);
\draw [black] (44.7,-26.22) -- (50.8,-24.58);
\fill [black] (50.8,-24.58) -- (49.9,-24.3) -- (50.16,-25.27);
\draw [black] (44.57,-28.14) -- (50.93,-30.76);
\fill [black] (50.93,-30.76) -- (50.38,-29.99) -- (50,-30.92);
\draw [black] (43.86,-29.18) -- (51.64,-37.42);
\fill [black] (51.64,-37.42) -- (51.45,-36.49) -- (50.73,-37.18);
\draw [black] (56.16,-22.08) -- (62.94,-17.32);
\fill [black] (62.94,-17.32) -- (62,-17.37) -- (62.58,-18.19);
\draw [black] (56.7,-23.8) -- (62.4,-23.8);
\fill [black] (62.4,-23.8) -- (61.6,-23.3) -- (61.6,-24.3);
\draw [black] (56.17,-25.51) -- (62.93,-30.19);
\fill [black] (62.93,-30.19) -- (62.56,-29.33) -- (61.99,-30.15);
\draw [black] (56.21,-33.55) -- (62.89,-37.95);
\fill [black] (62.89,-37.95) -- (62.5,-37.09) -- (61.95,-37.93);
\draw [black] (55.49,-26.21) -- (63.61,-37.19);
\fill [black] (63.61,-37.19) -- (63.54,-36.25) -- (62.74,-36.84);
\draw [black] (55.02,-26.49) -- (64.08,-44.91);
\fill [black] (64.08,-44.91) -- (64.17,-43.97) -- (63.27,-44.41);
\draw [black] (55.45,-29.46) -- (63.65,-18.04);
\fill [black] (63.65,-18.04) -- (62.78,-18.4) -- (63.59,-18.98);
\draw [black] (56.17,-30.19) -- (62.93,-25.51);
\fill [black] (62.93,-25.51) -- (61.99,-25.55) -- (62.56,-26.37);
\draw [black] (56.7,-31.9) -- (62.4,-31.9);
\fill [black] (62.4,-31.9) -- (61.6,-31.4) -- (61.6,-32.4);
\draw [black] (55.49,-34.31) -- (63.61,-45.19);
\fill [black] (63.61,-45.19) -- (63.53,-44.25) -- (62.73,-44.85);
\draw [black] (55.01,-36.9) -- (64.09,-18.3);
\fill [black] (64.09,-18.3) -- (63.29,-18.8) -- (64.18,-19.23);
\draw [black] (55.49,-37.19) -- (63.61,-26.21);
\fill [black] (63.61,-26.21) -- (62.74,-26.56) -- (63.54,-27.15);
\draw [black] (56.21,-37.95) -- (62.89,-33.55);
\fill [black] (62.89,-33.55) -- (61.95,-33.57) -- (62.5,-34.41);
\draw [black] (56.7,-39.6) -- (62.4,-39.6);
\fill [black] (62.4,-39.6) -- (61.6,-39.1) -- (61.6,-40.1);
\draw [black] (56.18,-41.29) -- (62.92,-45.91);
\fill [black] (62.92,-45.91) -- (62.55,-45.04) -- (61.98,-45.87);
\end{tikzpicture}

  }
  \caption[Sample autoencoder]{A possible neural architecture for an autoencoder with a 2-variable encoding layer. $W$ denotes weight matrices.}
  \label{fig:ae-example}
\end{figure}
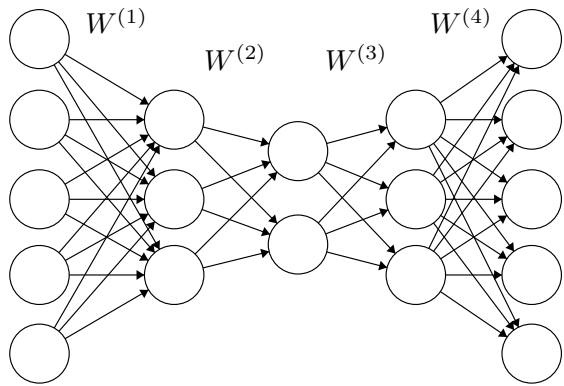

This structure is captured in a feedforward neural network. Since the objective is to reproduce the input data on the output layer, both $x$ and $r$ have the same dimension. $y$, however, can be higher-dimensional or lower-dimensional, depending on the properties desired. The AE can also have as many layers as needed, usually placed symmetrically in the encoder and decoder. Such a neural architecture can be observed in Fig.~\ref{fig:ae-example}.

In this case the encoder is made up of three layers, including the middle encoding one, while the decoder starts in the middle one and also spans three layers. 

\subsection{Activation functions of common use in autoencoders}
\begin{figure*}[htp!]
	\centering
	\begin{subfigure}[t]{0.3\textwidth}
		\centering
		\includegraphics[width=\linewidth]{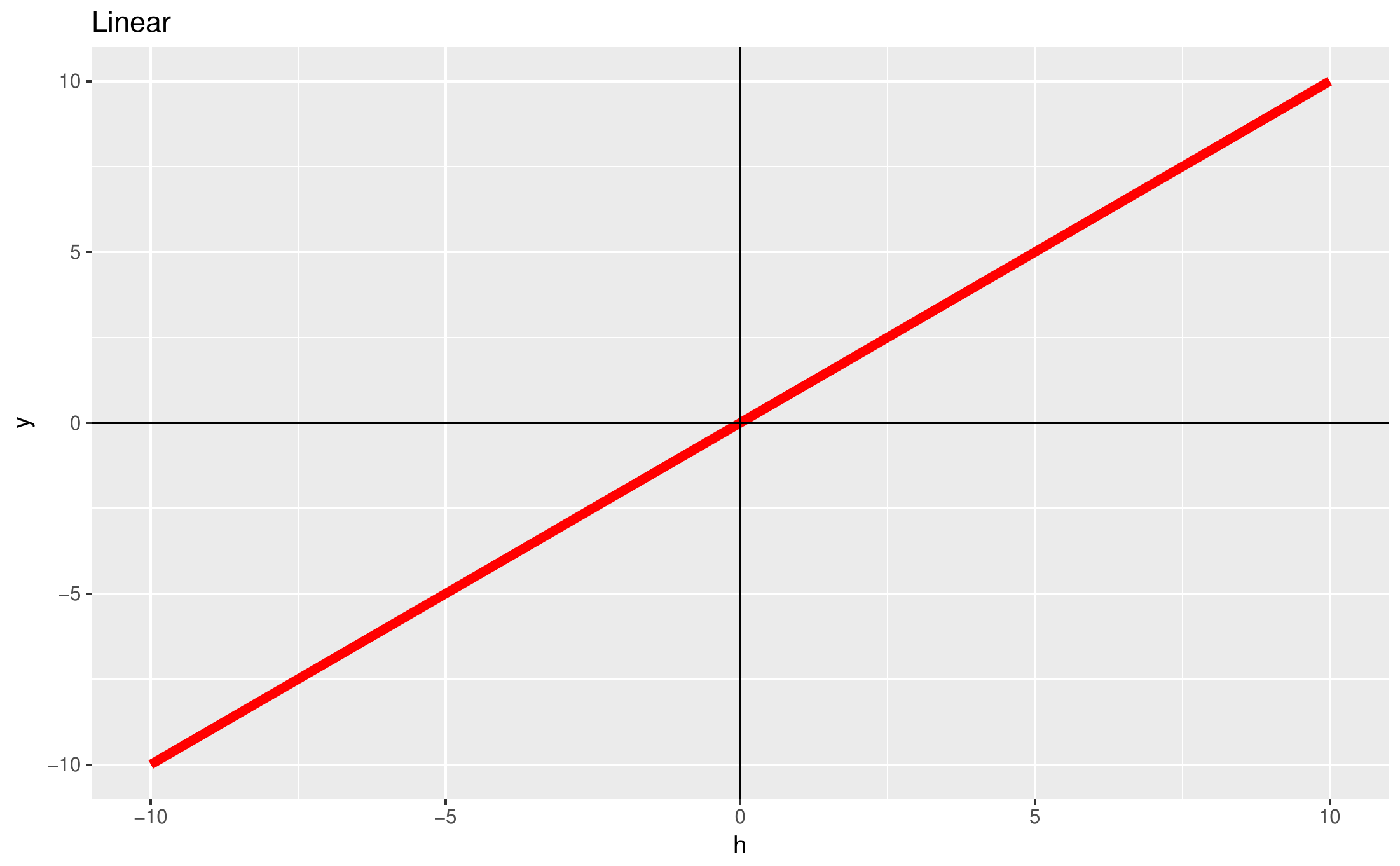} 
		\caption{Linear} \label{Fig.Linear}
	\end{subfigure}
	\begin{subfigure}[t]{0.3\textwidth}
		\centering
		\includegraphics[width=\linewidth]{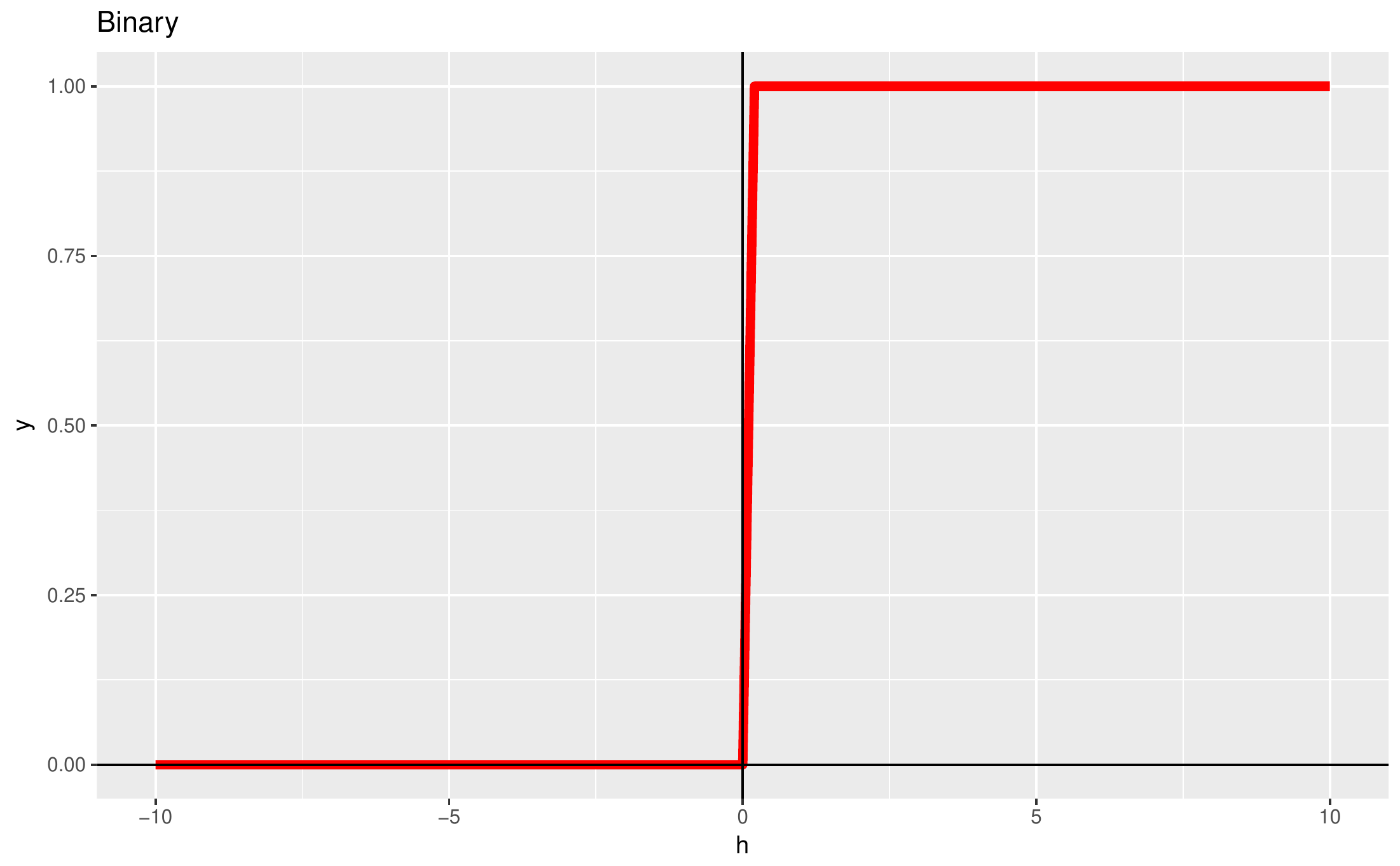} 
		\caption{Binary/Boolean} \label{Fig.Binary}
	\end{subfigure}
	\begin{subfigure}[t]{0.3\textwidth}
		\centering
		\includegraphics[width=\linewidth]{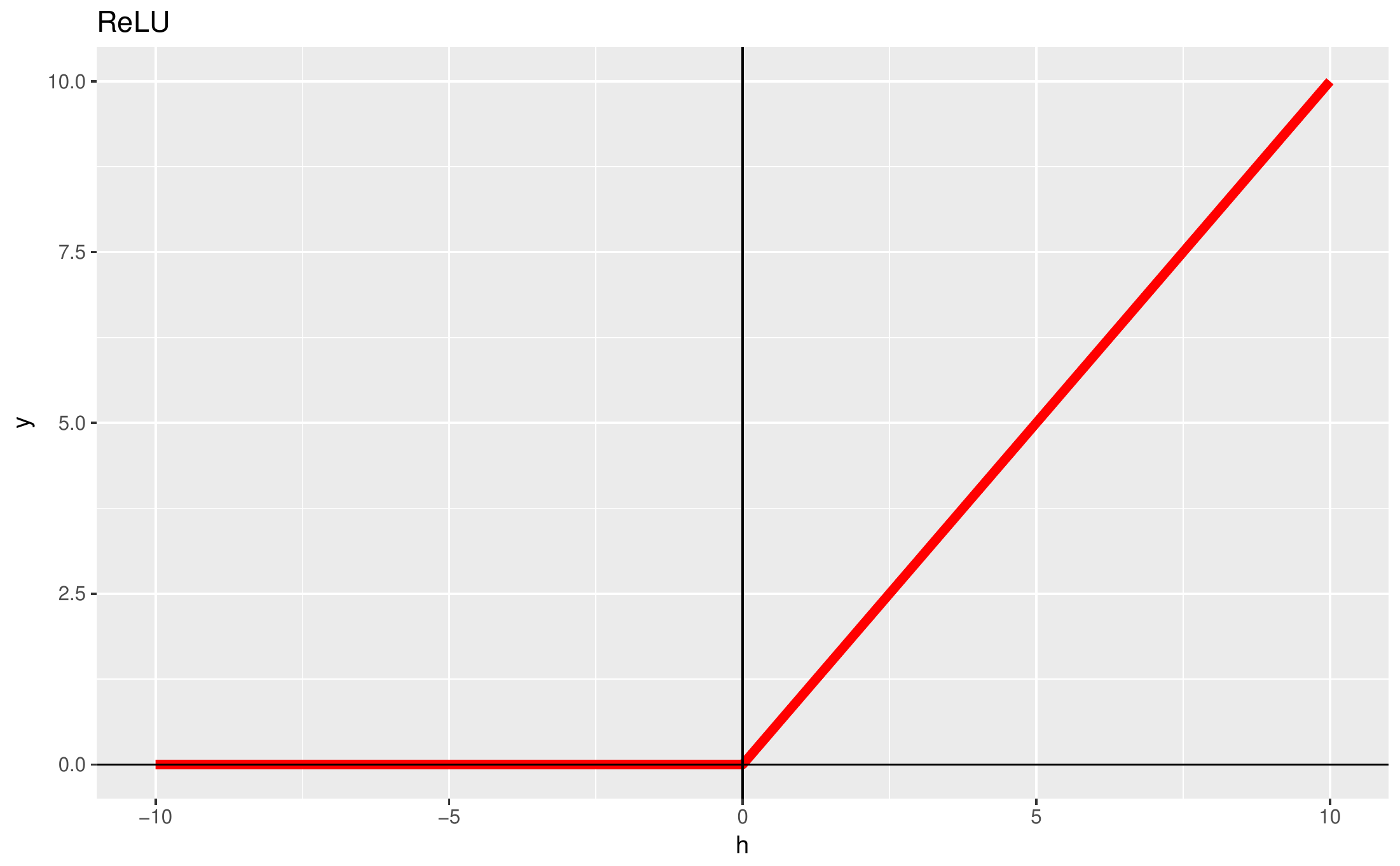} 
		\caption{ReLU} \label{Fig.ReLU}
	\end{subfigure}
	
	\begin{subfigure}[t]{0.3\textwidth}
		\centering
		\includegraphics[width=\linewidth]{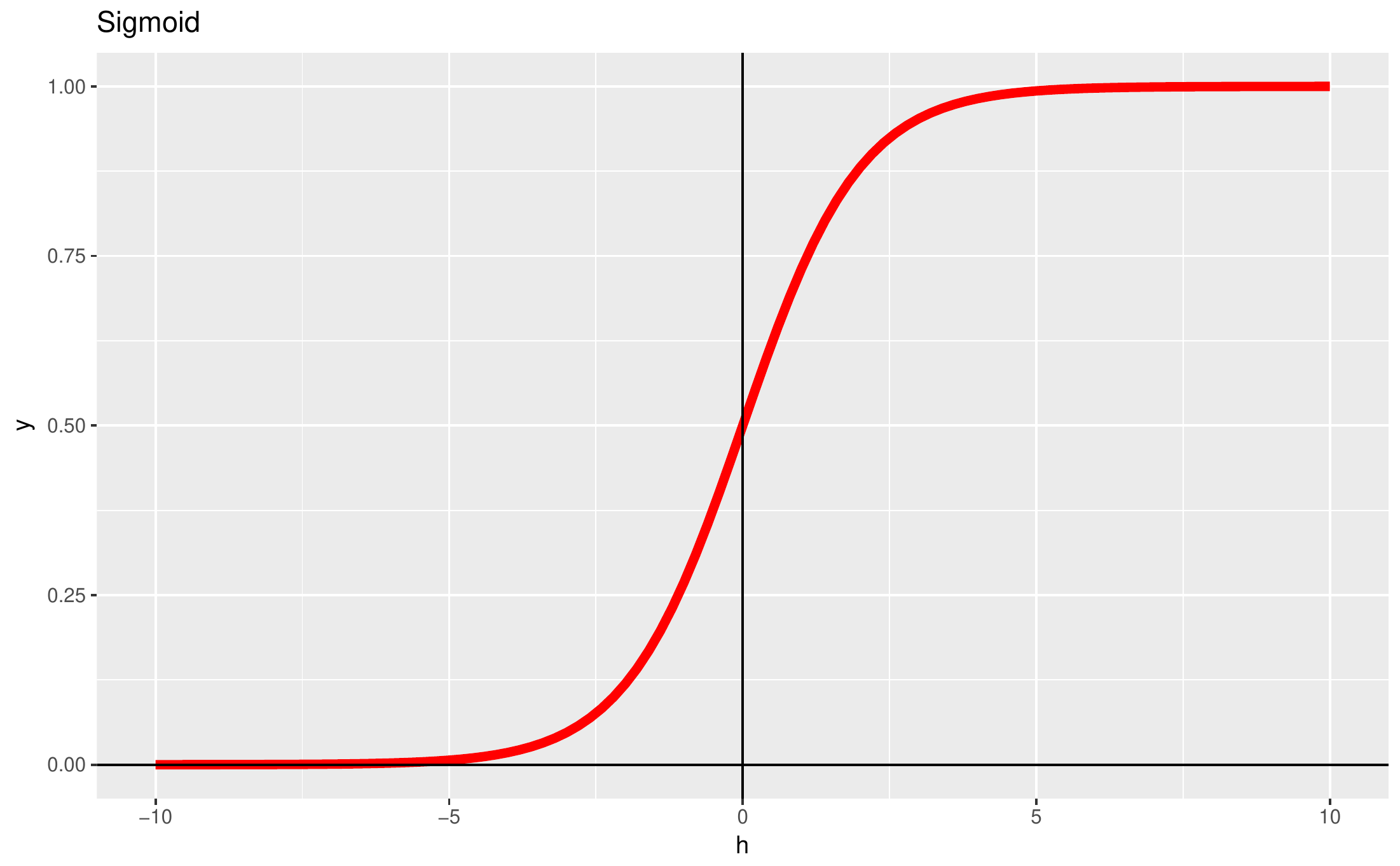} 
		\caption{Logistic} \label{Fig.Sigmoid}
	\end{subfigure}
	\begin{subfigure}[t]{0.3\textwidth}
		\centering
		\includegraphics[width=\linewidth]{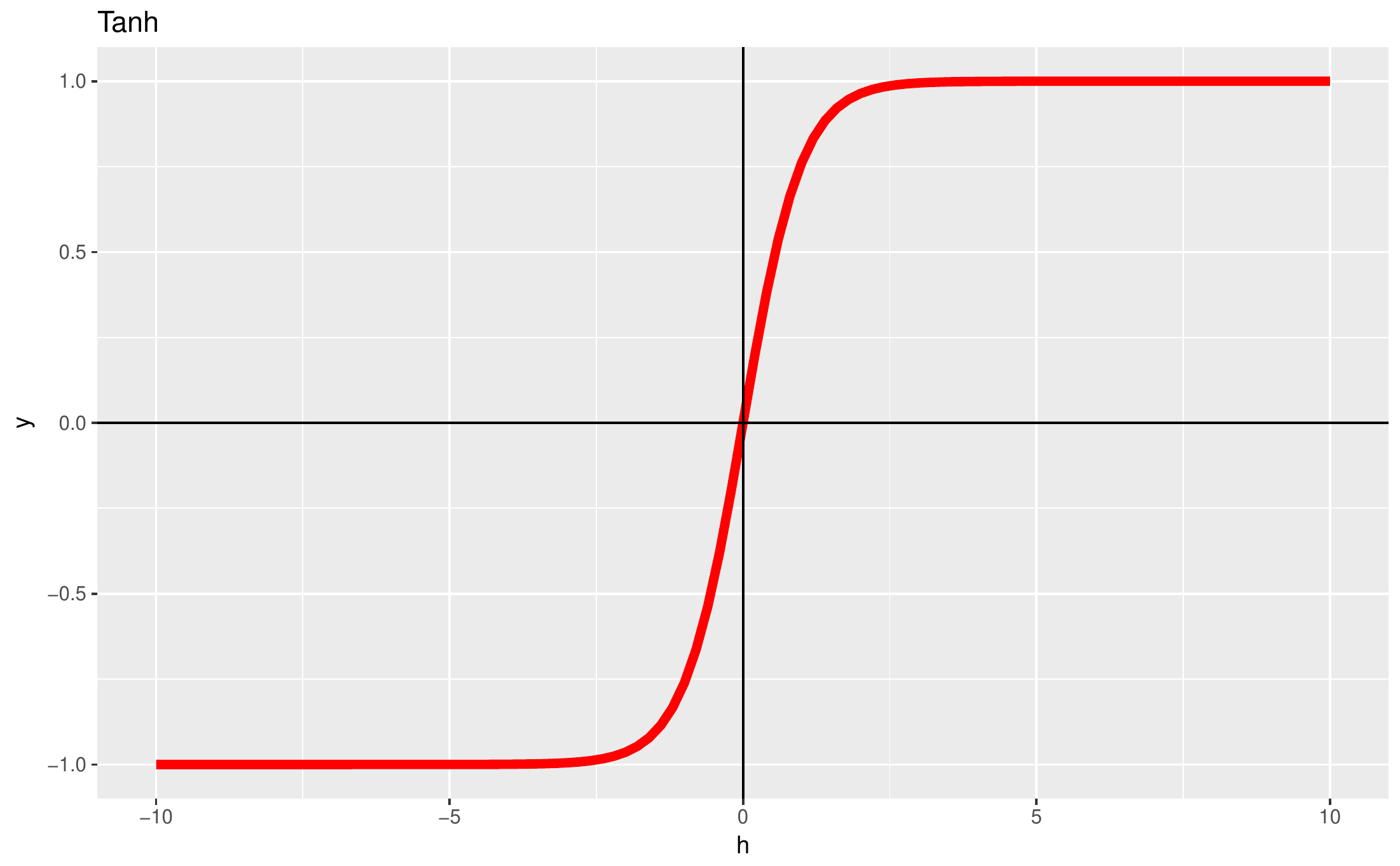} 
		\caption{Tanh} \label{Fig.Tanh}
	\end{subfigure}
	\begin{subfigure}[t]{0.3\textwidth}
		\centering
		\includegraphics[width=\linewidth]{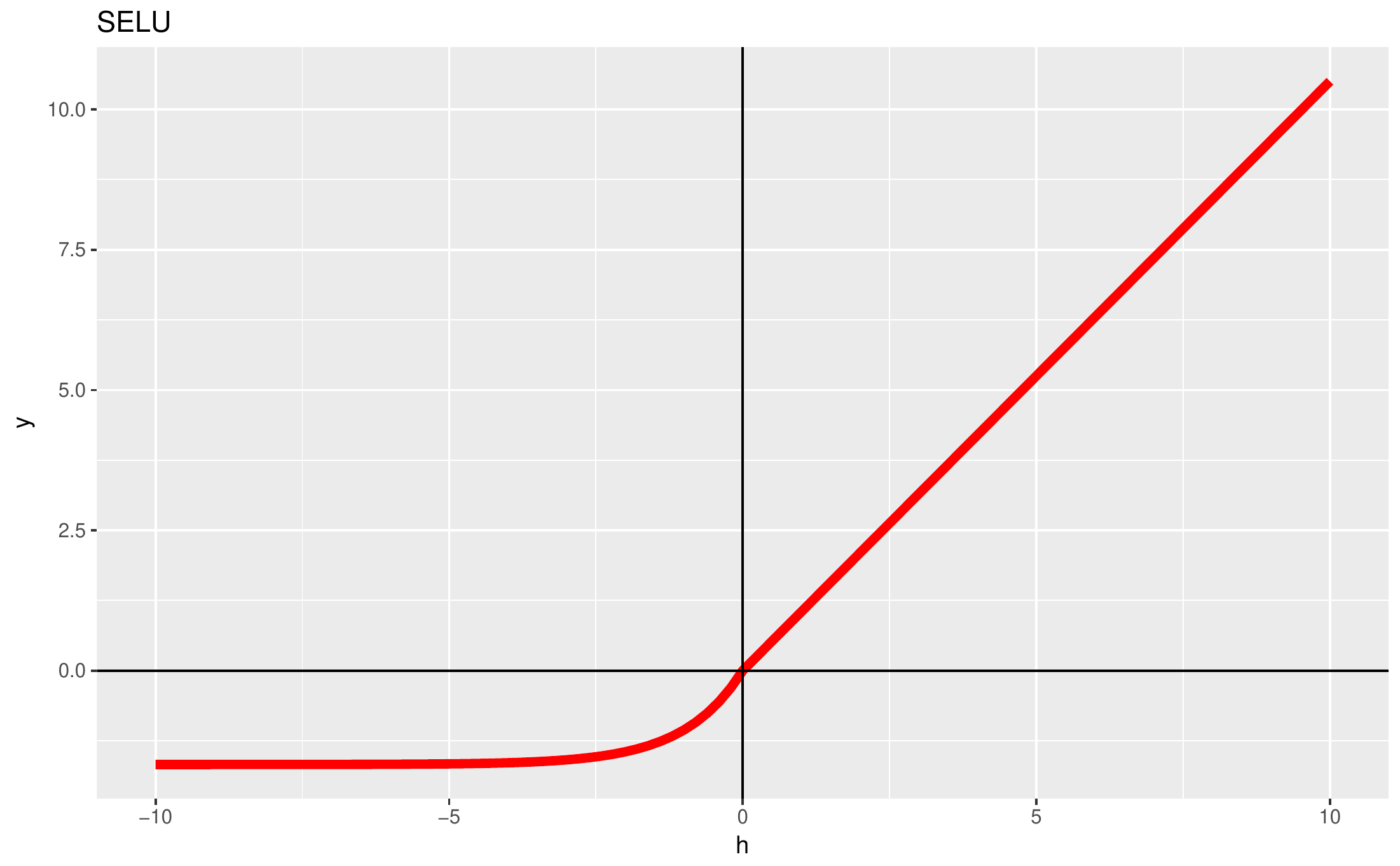} 
		\caption{SELU} \label{Fig.SELU}
	\end{subfigure}

	\caption{Common activation functions in ANNs.}
	\label{Fig.ActivationFunctions}
\end{figure*}

A unit located in any of the hidden layers of an ANN receives several inputs from the preceding layer. The unit computes the weighted sum of these inputs and eventually applies a certain operation, the so-called \textit{activation function}, to produce the output. The nonlinearity behavior of most ANNs is founded on the selection of the activation function to be used. Fig. \ref{Fig.ActivationFunctions} shows the graphical appearance of six of the most popular ones.

The activation functions shown in the first row are rarely used on AEs when it comes to learning higher level features, since they rarely provide useful representations. An undercomplete AE having one hidden layer made up of $k$ linear activation units (Eq.~\ref{Eq.linear}) and that minimizes the sum of squared errors is known to be equivalent to obtaining the $k$ principal components of the feature space via PCA \cite{ANNsPCA2,ANNsPCA,AutoencoderScoring}. AEs using binary/boolean activations (Eq.~\ref{Eq.binary}) \cite{deng_binary_2010,baldi_autoencoders_2012} are mostly adopted for educational uses, as McCulloch and Pitts \cite{McCullochPitts} cells are still used in this context. However, they also have some specific applications, such as data hashing as described in subsection \ref{Sect.Hashing}.

\begin{align}
  \label{Eq.linear} s_{\mathrm{linear}}(x) &= x \\
  \label{Eq.binary} s_{\mathrm{binary}}(x) &= [x > 0]
\end{align}
Note that square brackets denote Iverson's convention \cite{KnuthNotation} and evaluate to 0 or 1 according to the truthiness of the proposition. 

Rectified linear units (ReLU, Fig.~\ref{Fig.ReLU}, Eq.~\ref{Eq.relu}) are popular in many deep learning models, but it is an activation function that tends to degrade the AE performance. Since it always outputs 0 for negative inputs, it weakens the process of reconstructing the input features onto the outputs. Although they have been successfully used in \cite{RELUinAEs,RELUinAEs2}, the authors had to resort to a few detours. A recent alternative which combines the benefits of ReLU while circumventing these obstacles is the SELU function (\textit{Scaled Exponential Linear Units}, Fig.~\ref{Fig.SELU}, Eq.~\ref{Eq.selu}) \cite{SELU}. There are already some proposals of deep AEs based on SELU such as \cite{DeepAEwSELU}.
\begin{align}
  \label{Eq.relu} s_{\mathrm{relu}}(x) &= x[x > 0] \\
  \label{Eq.selu} s_{\mathrm{selu}}(x) &= \lambda \begin{cases} \alpha e^x-\alpha & x \leq 0 \\ x & x > 0 \end{cases},\text{ where }\lambda>1
\end{align}

Sigmoid functions are  undoubtedly the most common activations in AEs. The standard logistic function, popularly known simply as  sigmoid (Fig.~\ref{Fig.Sigmoid}, Eq.~\ref{Eq.sigm}), is probably the most frequently used. The hyperbolic tangent (Fig.~\ref{Fig.Tanh}, Eq.~\ref{Eq.tanh}) is also a sigmoid function, but it is symmetric about the origin and presents a steeper slope. According to LeCun \cite{EfficientBackprop} the latter should be preferred, since its derivative produces stronger gradients that the former.
\begin{align}
  \label{Eq.sigm} s_{\mathrm{sigm}}(x) &= \sigma(x) = \frac{1}{1+e^x} \\
  \label{Eq.tanh} s_{\mathrm{tanh}}(x) &= \tanh(x) = \frac{e^x-e^{-x}}{e^x+e^{-x}}
\end{align}

When designing AEs with multiple hidden layers, it is possible to use different activation functions in some of them. This would result in AEs combining the characteristics of several of these functions.

\subsection{Autoencoder groups according to network structure}

AEs could be grouped according to disparate principles, such as their structure, the learning algorithm they use, the loss function that guides the update of weights, their activation function or the field they are applied. In this section we focus on the first criterion, while the others will be further covered in following sections. 

As explained above, AEs are ANNs with a symmetrical structure. The decoder and the encoder have the same number of layers, with the number of units per layer in reverse order. The encoding layer is shared by both parts. Depending on the dimensionality of the encoding layer, AEs are said to be:

\begin{figure*}[h!]
	\centering
	\begin{subfigure}[t]{0.45\textwidth}
		\centering
		\includegraphics[width=0.7\linewidth]{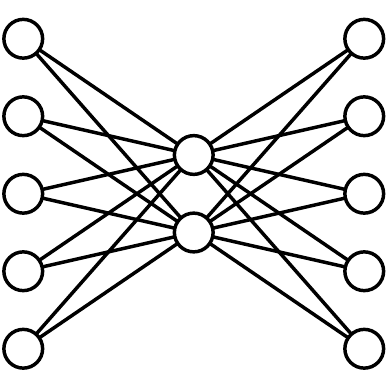} 
		\caption{Shallow undercomplete} \label{Fig.ShallowUnder}
	\end{subfigure}
	\hfill
	\begin{subfigure}[t]{0.45\textwidth}
		\centering
		\includegraphics[width=0.7\linewidth]{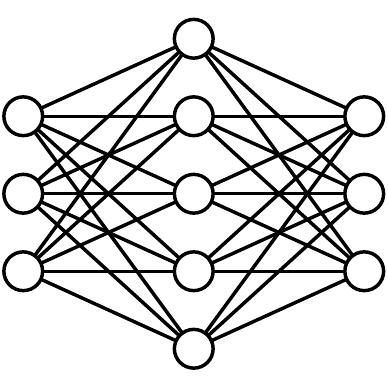} 
		\caption{Shallow overcomplete} \label{Fig.ShallowOver}
	\end{subfigure}
	
	\begin{subfigure}[t]{0.45\textwidth}
		\centering
		\includegraphics[width=0.7\linewidth]{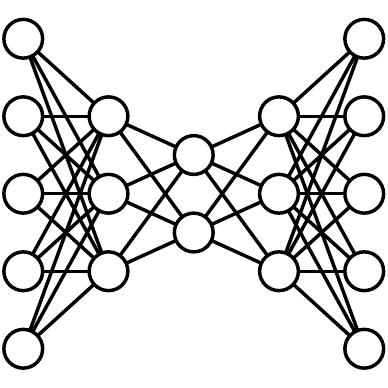} 
		\caption{Deep undercomplete} \label{Fig.DeepUnder}
	\end{subfigure}
	\hfill
	\begin{subfigure}[t]{0.45\textwidth}
		\centering
		\includegraphics[width=0.7\linewidth]{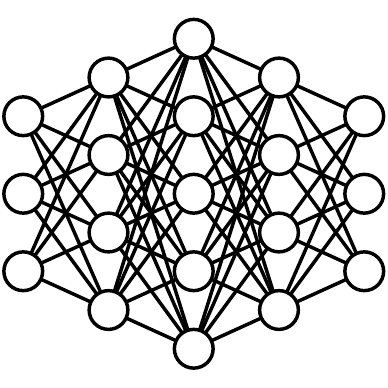} 
		\caption{Deep overcomplete} \label{Fig.DeepOver}
	\end{subfigure}
	\caption{Autoencoder models according to their structure.}
	\label{Fig.Structure}
\end{figure*}

\begin{itemize}
	\item \textit{Undercomplete}, if the encoding layer has a lower dimensionality than the input. The smaller number of units imposes a restriction, so during training the AE is forced to learn a more compact representation. This is achieved by fusing the original features according to the weights assigned through the learning process.
	
	\item \textit{Overcomplete}, otherwise. An encoding layer having the same or more units than the input could allow the AE to simply learn the identity function, copying the input onto the output. To avoid this behavior, usually other restrictions are applied as will be explained later.
\end{itemize}

Although the more popular AE configuration for dimensionality reduction is undercomplete, an overcomplete AE with the proper restrictions can also produce a compact encoding as explained in subsection \ref{Sec.SparseAE}.

In addition to the number of units per layer, the structure of an AE is also dependent of the number of layers. According to this factor, an AE can be:

\begin{itemize}
	\item \textit{Shallow}, when it only comprises three layers (input, encoding and output). It is the simplest AE model, since there is only one hidden layer (the encoding).
	
	\item \textit{Deep}, when it has more than one hidden layer. This kind of AE can be trained either layer by layer, as several shallow stacked AEs, or as a deep ANN \cite{DeepTrainingStrategies}.
\end{itemize}

These four types of AEs are visually summarized in Fig. \ref{Fig.Structure}. Shallow AEs are on the top row and deep ones in the bottom, while undercomplete AEs are on the left column and overcomplete on the right one.

\subsection{Autoencoder taxonomy}

\begin{figure*}[htp!]
	\centering
	\resizebox {.75\textwidth} {!} {
		
		\begin{tikzpicture}[
		font=\small,
		every node/.style = {shape=rectangle,
			draw, align=center},
		level 1/.style={sibling distance=10em},
		level 2/.style={grow=down, anchor=west, draw=gray, xshift=-1em,
			edge from parent path={([xshift=1em]\tikzparentnode.south west) |- (\tikzchildnode.west)}},
		level distance=5em,
		first/.style={level distance=6ex,xshift=-1em},
		second/.style={level distance=12ex,xshift=-1em},
		third/.style={level distance=18ex,xshift=-1em},
		],
		
		\node {Autoencoder taxonomy}
		child { node {Lower dimensionality}
			child[first]  { node {Basic} }
			child[second] { node {Convolutional} }
			child[third]  { node {LSTM-based} }
		}
		child { node {Regularization}
			child[first]  { node {Sparse} }
			child[second] { node {Contractive} }
		}
		child { node {Noise tolerance}
			child[first]  { node {Denoising} }
			child[second] { node {Robust} }
		}
		child { node {Generative model}
			child[first]  { node {Variational} }
			child[second] { node {Adversarial} }
		};
		\end{tikzpicture}
	}
	\caption{Taxonomy: most popular autoencoders classified according to the charasteristics they induce in their encodings}
	\label{fig:autoencoder-taxonomy}
\end{figure*}
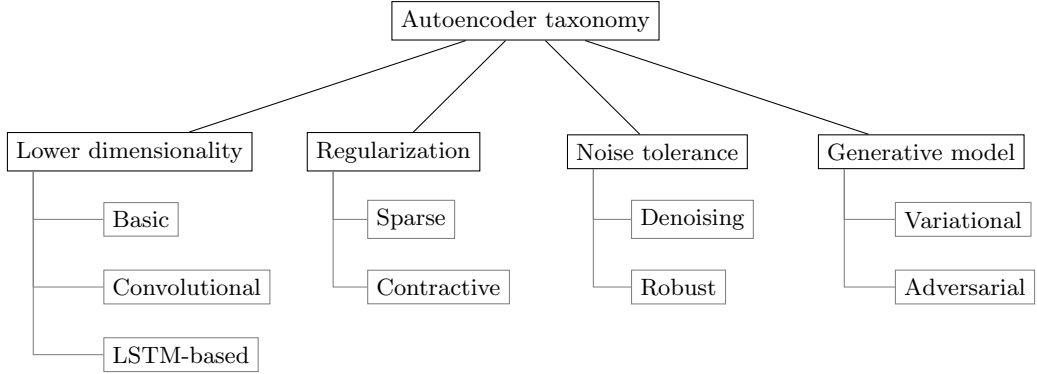

As stated before, a taxonomy of AEs can be built according to different criteria. Here the interest is mainly on the properties of the inferred model regarding the feature fusion task. Conforming to this principle, we have elaborated the taxonomy shown in Fig. \ref{fig:autoencoder-taxonomy}. As can be seen, there are four main categories in this taxonomy:

\paragraph{Lower dimensionality}
High-dimensional data can be an issue when using most classifiers and especially shallow neural networks, since they do not perform any kind of high-level feature learning and are then forced to optimize a notable amount of parameters. This task may be eased by just lowering the dimensionality of the data, and this is the aim of the basic AE, which is thoroughly explained in Section~\ref{Sec.BasicAE}. Decreasing the dimensionality of specific types of data, such as images or sequences, can be treated by domain specific AEs detailed in Section~\ref{Sec.DomainSpecific}.

\paragraph{Regularization}
Sometimes, learned features are required to present special mathematical properties. AEs can be easily modified in order to reach encodings that verify them. The main regularizations that can be applied to AEs are portrayed in Section~\ref{Sec.Regularization}.

\paragraph{Noise tolerance}
In addition to different properties, a desirable trait for the encoded space may be robustness in the face of noisy data. Two distinct approaches to this problem using AEs are gathered in Section~\ref{Sec.NoiseTolerance}.

\paragraph{Generative model}
The transformation from the original feature space onto the encoded space may not be the main objective of an AE. Occasionally it will be useful to map new samples in the encoded space onto the original features. In this case, a generative model is needed. Those based in AEs are specified in Section~\ref{Sec.Generative}.

\subsection{Usual applications}
The term autoencoder is very broad, referring to multiple learning models based on both fully-connected feed-forward ANNs and other types of ANNs, and even models completely unrelated to that structure. Similarly, the application fields of AEs are also varied.  In this work we pay attention specifically to AEs whose basic model is that of an ANN. In addition, we are especially interested in those whose objective is the fusion of characteristics by means of nonlinear techniques.

Reducing the dimensionality of a feature space using AEs can be achieved following disparate approaches. Most of them are reviewed in Section \ref{Sect.AEforFF}, starting with the basic AE model, then advancing to those that include a regularization, that present noise tolerance, etc. The goal is to provide a broad view of the techniques that AEs rely on to perform feature fusion.

Besides feature extraction, which is our main focus, there are AE models designed for other applications such as outlier detection, hashing, data compression or data generation. In Sections~\ref{Sec.Generative} and \ref{Sec.OtherAEs} some of these AEs will be briefly portrayed, and in Section \ref{Sec.Applications} many of their applications will be shortly reviewed.

\section{Autoencoders for feature fusion}\label{Sect.AEforFF}

As has been already established, AEs are tools originally designed for finding useful representations of data by learning nonlinear ways to combine their features. Usually, this leads to a lower-dimensional space, but different modifications can be applied in order to discover features which satisfy certain requirements. All of these possibilities are discussed in this section, which begins by establishing the foundations of the most basic AE, and later encompasses several diverse variants, following the proposed taxonomy: those that provide regularizations are followed by AEs presenting noise tolerance, generative models are explained afterwards, then some domain specific AEs and finally two variations which do not fit into any previous category.

\subsection{Basic autoencoder}\label{Sec.BasicAE}

The main objective of most AEs is to perform a feature fusion process where learned features present some desired traits, such as lower dimensionality, higher sparsity or desirable analytical properties. The resulting model is able to map new instances onto the latent feature space. All AEs thus share a common origin, which may be called the basic AE \cite{bourlard_auto-association_1988}.

The following subsections define the structure of a basic AE, establish their objective function, describe the training process while enumerating the necessary algorithms for this task, and depict how a deep AE can be initialized by stacking several shallow ones.

\subsubsection{Structure}

The structure of a basic AE, as shown in the previous section, is that of a feed forward ANN where layers are of symmetrical amount of units. Layers need not be symmetrical in the sense of activation functions or weight matrices.

The simplest AE consists of just one hidden layer, and is defined by two weight matrices and two bias vectors:
\begin{align}
  y&=f(x)=s_1(W^{(1)}x+b^{(1)}),\\
  r&=g(y)=s_2(W^{(2)}y+b^{(2)}),
\end{align}
where $s_1$ and $s_2$ denote the activation functions, which usually are nonlinear.

Deep AEs are the natural extension of this definition to a higher number of layers. We will call the composition of functions in the encoder $f$, and the composition of functions in the decoder $g$.

\subsubsection{Objective function}

AEs generally base their objective function on a per-instance loss function $\mathcal L:\mathbb R^d\times \mathbb R^d\rightarrow \mathbb R$:

\begin{equation}
\mathcal J(W,b;S)= \sum_{x \in S} \mathcal L(x, (g\circ f)(x))
\end{equation}
where $f$ and $g$ are the encoding and decoding functions determined by the weights $W$ and biases $b$, assuming activation functions are fixed, and $S$ is a set of samples. The objective of an AE is thus to optimize $W$ and $b$ in order to minimize $\mathcal J$.

For example, a typical loss function is the mean squared error (MSE):
\begin{equation}
\mathcal L_{\mathrm{MSE}}(u, v)=\left\lVert u - v\right\rVert_2^2~.
\end{equation}
Notice that multiplying by constants or performing the square root of the error induced by each instance does not alter the process, since these operations preserve numerical order. As a consequence, the root mean squared error (RMSE) is an equivalent loss metric.

When a probabilistic model is assumed for the input samples, the loss function is chosen as the negative log-likelihood for the example $x$ given the output $(g\circ f)(x)$ \cite{LayerwiseTraining}. For instance, when input values are binary or modeled as bits, cross-entropy is usually the preferred alternative for the loss function:
\begin{equation}
  \mathcal L_{\mathrm{CE}}(u, v)=-\sum_{k=1}^d u_k \log v_k + (1 - u_k)\log(1 - v_k)~.
\end{equation}

\subsubsection{Training}\label{Sec.Training}

Usual algorithms for optimizing weights and biases in AEs are stochastic gradient descent (SGD) \cite{robbins1951stochastic} and some of its variants, such as AdaGrad \cite{duchi2011adaptive}, RMSProp \cite{tieleman2012lecture} and Adam \cite{kingma2015adam}. Other applicable algorithms which are not based on SGD are L-BFGS and conjugate gradient \cite{ngiam2011optimization}.

The foundation of these algorithms is the technique of gradient descent \cite{cauchy1847methode}. Intuitively, at each step, the gradient of the objective function with respect to the parameters shows the direction of steepest slope, and allows the algorithm to modify the parameters in order to search for a minimum of the function. 

In order to compute the necessary gradients, the backpropagation algorithm \cite{Backpropagation} is applied. Backpropagation performs this computation by calculating several intermediate terms from the last layer to the first.

AEs, like many other machine learning techniques, are susceptible to overfitting of the training data. To avoid this issue, a regularization term can be added to the objective function which causes a \textit{weight decay} \cite{krogh1992wd}. This improves the generalization ability and encourages smaller weights that produce good reconstructions. Weight decay can be introduced in several ways, but essentially consists in a term depending on weight sizes that will attempt to limit their growth. For example, the resulting objetive function could be
\begin{equation}
  \label{Eq.wd}
  \mathcal J(W,b;S)=\sum_{x\in S}\mathcal L(x, (g\circ f)(x)) + \lambda \sum_{i}w_i^2
\end{equation}
where $w_i$ traverses all the weights in $W$ and $\lambda$ is a parameter determining the magnitude of the decay.

Further restrictions and regularizations can be applied. A specific constraint that can be imposed is to tie the weight matrices symmetrically, that is, in a shallow AE, to set $W^{(1)}=(W^{(2)})^T$, and the natural extension to deep AEs. This allows to optimize a lower amount of parameters, so the AE can be trained faster, while maintaining the desired architecture.

\subsubsection{Stacking}

When AEs are deep, the success of the training process relies heavily on a good weight initialization, since there can be from tens to hundreds of thousands of them. This weight initialization can be performed by \textit{stacking} successive shallow AEs \cite{LayerwiseTraining}, that is, training the AE layer by layer in a greedy fashion.

\begin{figure}[ht!]
  \centering
  \includegraphics{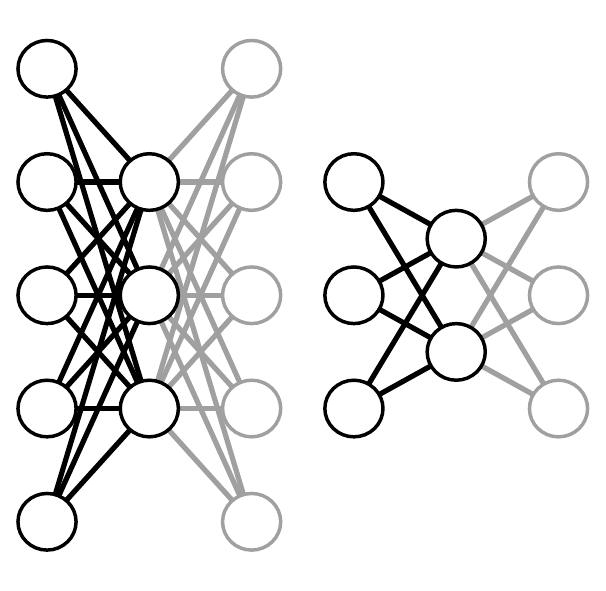}
  \caption{\label{Fig.StackedAutoencoder}Greedy layer-wise training of a deep AE with the architecture shown in Fig~\ref{Fig.DeepUnder}. Units drawn in black designate layers of the final AE, and gray ones indicate layers that are not part of the unrolled AE during the fine-tuning phase.}
\end{figure}

The training process begins by training only the first hidden layer as the encoding of a shallow AE, as shown by the network on the left of Fig.~\ref{Fig.StackedAutoencoder}. After this step, the second hidden layer is trained, using the first hidden layer as input layer, as displayed on the right. Inputs are computed via a forward pass of the original inputs through the first layer, with the weights determined during the previous stage. Each successive layer up to the encoding is trained the same way.

After this layer-wise training, initial weights for all layers preceding and including the encoding layer will have been computed. The AE is now ``unrolled'', i.e. the rest of layers are added symetrically with weight matrices resulting from transposing the ones from each corresponding layer. For instance, for the AE trained in Fig.~\ref{Fig.StackedAutoencoder}, the unrolled AE would have the structure shown in Fig.~\ref{Fig.DeepUnder}.

Finally, a fine-tuning phase can be performed, optimizing the weights by backpropagating gradients through the complete structure with training data.

\subsection{Regularization}\label{Sec.Regularization}

Encodings produced by basic AEs do not generally present any special properties. When learned features are required to verify some desirable traits, some regularizations may be achieved by adding a penalization for certain behaviors $\Omega$ to the objetive function: 
\begin{equation}
  \mathcal J(W,b;S)=\sum_{x\in S} \mathcal L(x, (g\circ f)(x)) + \lambda\Omega(W,b;S)~.
\end{equation}

\subsubsection{Sparse autoencoder}\label{Sec.SparseAE}

Sparsity in a representation means most values for a given sample are zero or close to zero. Sparse representations are resembling of the behavior of simple cells in the mammalian primary visual cortex, which is believed to have evolved to discover efficient coding strategies \cite{olshausen1997sparse}. This motivates the use of transformations of samples into sparse codes in machine learning. A model of sparse coding based on this behavior was first proposed in \cite{olshausen1996emergence}.

Sparse codes can also be overcomplete and meaningful. This was not necessarily the case in basic AEs, where an overcomplete code would be trained to just copy inputs onto outputs.

When sparsity is desired in the encoding generated by an AE, activations of the encoding layer need to have low values in average, which means units in the hidden layer usually do not fire. The activation function used in those units will determine this low value: in the case of sigmoid and ReLU activations, low values will be close to 0; this value will be -1 in the case of $\tanh$, and $-\lambda\alpha$ in the case of a SELU. 

The common way to introduce sparsity in an AE is to add a penalty to the loss function, as proposed in \cite{lee_sparse_2008} for Deep Belief Networks. In order to compare the desired activations for a given unit to the actual ones, these can be modeled as a Bernoulli random variable, assuming a unit can only either fire or not. For a specific input $x$, let
\begin{equation}
  \hat\rho_i=\frac{1}{\left\lvert S\right\rvert} \sum_{x\in S} f_i(x)
\end{equation}
be the average activation value of an unit in the hidden layer, where $f=(f_1,f_2,\dots f_c)$ and $c$ is the number of units in the encoding. $\hat\rho_i$ will be the mean of the associated Bernoulli distribution.

\begin{figure}[ht!]
	\centering
	\includegraphics[width=0.9\linewidth]{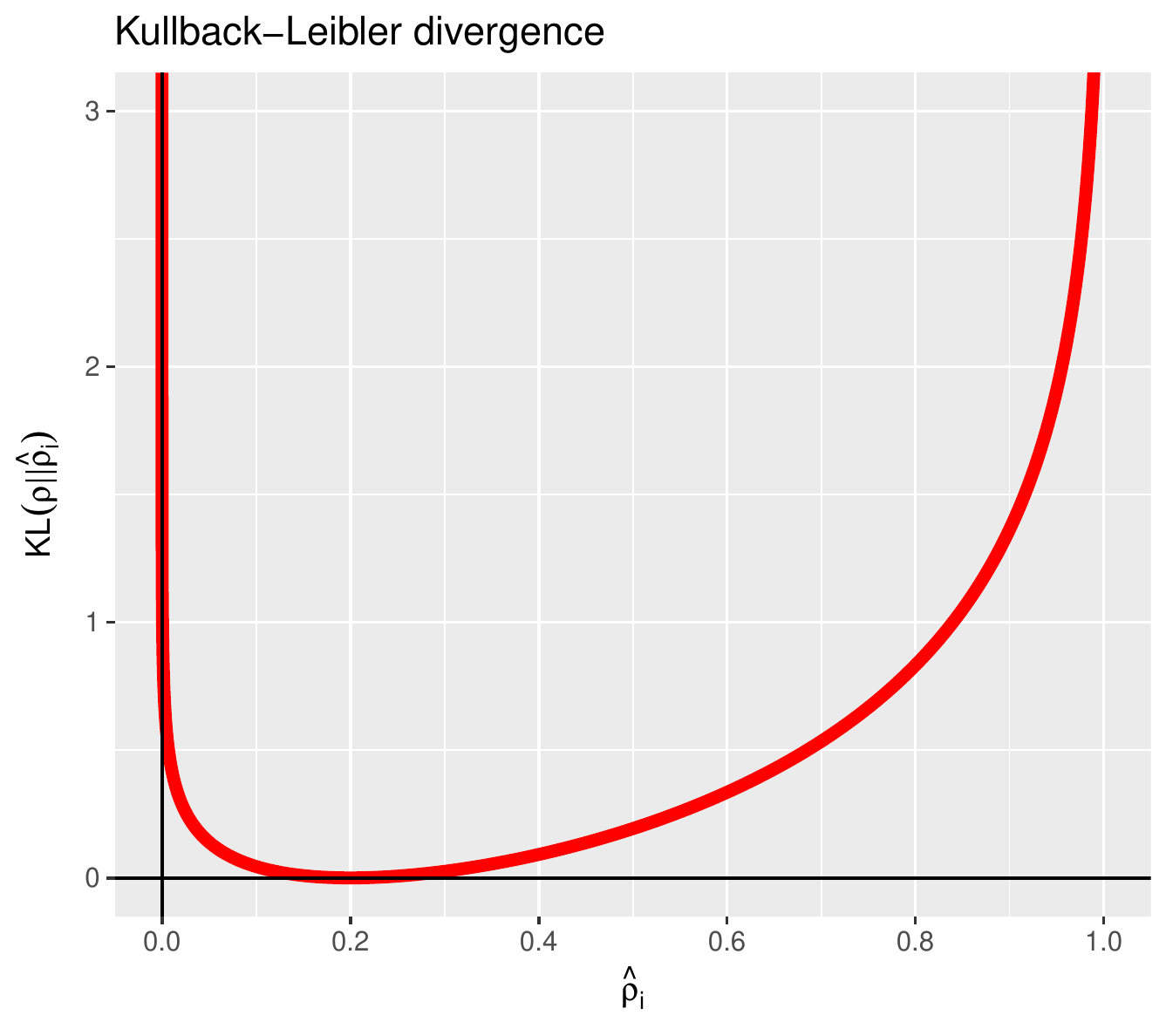} 

	\caption{Values of Kullback-Leibler divergence for a unit with average activation $\hat\rho_i$.}
	\label{Fig.KLdivergence}
\end{figure}

Let $\rho$ be the desired average activation. The Kullback-Leibler divergence \cite{kullback1951information} between the random variable defined by unit $i$ and the one corresponding to the desired activations will measure how different both distributions are \cite{ng2011sparse}:
\begin{equation}\label{Eq.KLdivergence}
  \mathrm{KL}(\rho\Vert \hat\rho_i)=\rho \log\frac\rho{\hat\rho_i} + (1 - \rho)\log\frac{1 - \rho}{1 - \hat\rho_i}~.
\end{equation}
Fig.~\ref{Fig.KLdivergence} shows the penalty caused by Kullback-Leibler divergence for a hidden unit when the desired average activation is $\rho=0.2$. Notice that the penalty is very low when the average activation is near the desired one, but grows rapidly as it moves away and tends to infinity at 0 and 1.

The resulting penalization term for the objective function is
\begin{equation}
  \Omega_{\mathrm{SAE}}(W,b;S)=\sum_{i=1}^c \mathrm{KL}(\rho\Vert \hat\rho_i)~,
\end{equation}
where the average activation value $\hat\rho_i$ depends on the parameters of the encoder and the training set $S$.

There are other modifications that can lead an encoder-decoder architecture to produce a sparse code. For example, applying a sparsifying logistic activation function in the encoding layer of a similar energy-based model, which forces a low activation average \cite{SparsifyingLogistic}, or using a Sparse Encoding Symmetric Machine \cite{SESM} which optimizes a loss function with a different sparsifying penalty.

\subsubsection{Contractive autoencoder}

High sensitivity to perturbations in input samples could lead an AE to generate very different encodings. This is usually inconvenient, which is the motivation behind the contractive AE. It achieves local invariance to changes in many directions around the training samples, and is able to more easily discover lower-dimensional manifold structures in the data.

Sensitivity for small changes in the input can be measured as the Frobenius norm $\lVert\cdot\rVert_F$ of the Jacobian matrix of the encoder $J_f$:
\begin{equation}
  \left\lVert J_f(x) \right\rVert_F^2=
  \sum_{j=1}^d\sum_{i=1}^c \left(\frac{\partial f_i}{\partial x_j}\left(x\right)\right)^2~.
\end{equation}
The higher this value is, the more unstable the encodings will be to perturbations on the inputs.

A regularization is built from this measure into the objective function of the contractive AE:
\begin{equation}
  \Omega_{\mathrm{CAE}}(W,b;S) = \sum_{x\in S}\left\lVert J_f(x) \right\rVert_F^2~.
\end{equation}

A particular case of this induced contraction is the usage of L2 weight decay with a linear encoder: in this situation, the only way to produce a contraction is to maintain small weights. In the nonlinear case, however, contraction can be encouraged by pushing hidden units to the saturated region of the activation function.

The contractive AE can be sampled \cite{generativeCAE}, that is, it can generate new instances from the learned model, by using the Jacobian of the encoder to add a small noise to another point and computing its codification. Intuitively, this can be seen as moving small steps along the tangent plane defined by the encoder in a point on the manifold modeled.

\subsection{Noise tolerance}\label{Sec.NoiseTolerance}

A standard AE can learn a latent feature space from a set of samples, but it does not guarantee stability in the presence of noisy instances, nor it is able to remove noise when reconstructing new samples. In this section, two variants that tackle this problem are discussed: denoising and robust AEs.

\subsubsection{Denoising autoencoder}

A denoising AE or DAE \cite{vincent_denoising_2008} learns to generate robust features from inputs by reconstructing partially destroyed samples. The use of AEs for denoising had been introduced earlier \cite{yann1987modeles}, but this technique leverages the denoising ability of the AE to build a latent feature space which is more resistant to corrupted inputs, thus its applications are broader than just denoising. 

\begin{figure}[ht!]
	\centering
	\includegraphics[width=0.7\linewidth]{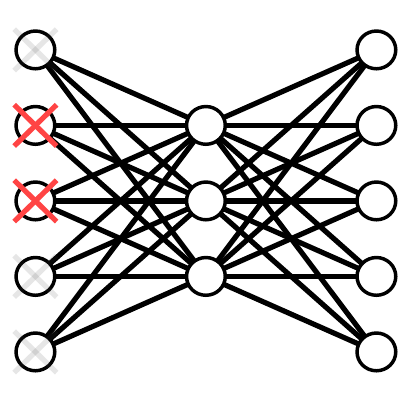} 

	\caption{Illustration of the training phase of a denoising AE. For each input sample, some of its components are randomly selected and set to 0, but the reconstruction error is computed by comparing to the original, non-corrupted data.}
	\label{Fig.Denoising}
\end{figure}

The structure and parameters of a denoising AE are identical to those of a basic AE. The difference here lies in a stochastic corruption of the inputs which is applied during the training phase. The corrupting technique proposed in \cite{vincent_denoising_2008}, as illustrated by Fig.~\ref{Fig.Denoising}, is to randomly choose a fixed amount of features for each training sample and set them to 0. The reconstructions of the AE are however compared to the original, uncorrupted inputs. The AE will be thus be trained to guess the missing values.

Formally, let $q(\tilde x\vert x)$ be a stochastic mapping performing the partial destruction of values described above, the denoising AE recieves $\tilde x\sim q(\tilde x\vert x)$ as input and minimizes
\begin{equation}
  \mathcal J_{\mathrm{DAE}}(W,b;S)=\sum_{x\in S}\mathbb E_{\tilde x \sim q(\tilde x \vert x)} \left[\mathcal L(x, (g\circ f)(\tilde x))\right]~.
\end{equation}

A denoising AE does not need further restrictions or regularizations in order to learn a meaningful coding from the data, which means it can be overcomplete if desired. When it has more than one hidden layer, it can be trained layer-wise. For this to be done, uncorrupted inputs are computed as outputs of the previous layers, these are then corrupted and provided to the network. Note that after the denoising AE is trained, it is used to compute higher-level representations without corrupting the input data.

The training technique allows for other possible corruption processes, apart from forcing some values to 0 \cite{vincent2010stacked}. For instance, additive Gaussian noise
\begin{equation}
  \tilde x \sim \mathcal N(x, \sigma^2 \mathrm I)~,
\end{equation}
which randomly offsets each component of $x$ with the same variance, or \textit{salt-and-pepper} noise, which sets a fraction of the elements of the input to their minimum or maximum value, according to a uniform distribution.

\subsubsection{Robust autoencoder}

Training an AE to recover from corrupted data is not the only way to induce noise tolerance in the generated model. An alternative is to modify the loss function used to minimize the reconstruction error in order to dampen the sensitivity to different types of noise.

Robust stacked AEs \cite{qi_robust_2014} apply this idea, and manage to be less affected by non-Gaussian noise than standard AEs. They achieve this by using a different loss function based on \textit{correntropy}, a localized similarity measure defined in \cite{liu_correntropy_2006}. 

\begin{align}
  \mathcal L_{\mathrm{MCC}}(u, v)&=-\sum_{k=1}^d\mathcal K_{\sigma}(u_k-v_k),\\\text{where }
  \mathcal K_{\sigma}(\alpha)&=\frac{1}{\sqrt{2\pi}\sigma}\exp\left(-\frac{\alpha^2}{2\sigma^2}\right),
\end{align}
and $\sigma$ is a parameter for the kernel $\mathcal K$.

Correntropy specifically measures the probability density that two events are equal. An advantage of this metric is it being less affected by outliers than MSE. Robust AEs attempt to maximize this measure (equivalently, minimize negative correntropy), which translates in a higher resilience to non-Gaussian noise.

\subsection{Domain specific autoencoders}\label{Sec.DomainSpecific}

The following two AEs are based on the standard type, but are designed to model very specific kinds of data, such as images and sequences.

\paragraph{Convolutional autoencoder \cite{ConvolutionalAE}} Standard AEs do not explicitly consider the 2-dimensional structure when processing image data. Convolutional AEs solve this by making use of convolutional layers instead of fully connected ones. In these, a global weight matrix is used and the convolution operation is applied in order to forward pass values from one layer to the next. The same matrix is flipped over both dimensions and used for the reconstruction phase. Convolutional AEs can also be stacked and used to initialize CNNs \cite{CNNsLeCun}, which are able to perform classification of images.

\paragraph{LSTM autoencoder \cite{LSTMAE}} A basic AE is not designed to model sequential data, an LSTM AE achieves this by placing Long-Short-Term Memory (LSTM) \cite{LSTM} units as encoder and decoder of the network. The encoder LSTM reads and compresses a sequence into a fixed-size representation, from which the decoder attempts to extract the original sequence in inverse order. This is especially useful when data is sequential and large, for example video data. A further possible task is to predict the future of the sequence from the representation, which can be achieved by attaching an additional decoder trained for this purpose.

\subsection{Generative models}\label{Sec.Generative}

In addition to the models already described, which essentially provide different mechanisms to reduce the dimensionality of a set of variables, the following ones also produce a generative model from the training data. Generative models learn a distribution in order to be able to draw new samples, different from those observed. AEs can generally reconstruct encoded data, but are not necessarily able to build meaningful outputs from arbitrary encodings. Variational and adversarial AEs learn a model of the data from which new instances can be generated.

\paragraph{Variational autoencoder \cite{VariationalAE}} This kind of AE applies a variational Bayesian \cite{VariationalBayes} approach to encoding. It assumes that a latent, unobserved random variable $\mathbf{y}$ exists, which by some random process leads to the observations, $\mathbf{x}$. Its objective is thus to approximate the distribution of the latent variable given the observations. Variational AEs replace deterministic functions in the encoder and decoder by stochastic mappings, and compute the objective function in virtue of the density functions of the random variables:
\begin{multline}
  \mathcal L_{\mathrm{VAE}}(\theta, \phi; \mathbf{x})=\\\mathrm{KL}(
  q_{\phi}(\mathbf{y}\vert\mathbf{x})
  \Vert
  p_{\theta}(\mathbf{y})
  )
  -
  \mathbb E_{  q_{\phi}(\mathbf{y}\vert\mathbf{x})}\left[\log p_{\theta}(\mathbf{x}\vert\mathbf{y})\right]~,
\end{multline}
where $q$ is the distribution approximating the true latent distribution of $\mathbf{y}$, and $\theta,\phi$ are the parameters of each distribution. Since variational AEs allow sampling from the learned distribution, applications usually involve generating new instances \cite{VAEgenerating,rezende2014stochastic}.

\paragraph{Adversarial autoencoder \cite{AdversarialAE}} It brings the concept of Generative Adversarial Networks \cite{GAN} to the field of AEs. It models the encoding by imposing a prior distribution, then training a standard AE and, concurrently, a discriminative network trying to distinguish codifications from samples from the imposed prior. Since the generator (the encoder) is trained to fool the discriminator as well, encodings will tend to follow the imposed distribution. Therefore, adversarial AEs are also able generate new meaningful samples. \\

Other generative models based on similar principles are Variational Recurrent AEs \cite{VRAE}, PixelGAN AEs \cite{PixelGAN} and Adversarial Symmetric Variational AEs \cite{ASVAE}.

\subsection{Other autoencoders farther from feature fusion}\label{Sec.OtherAEs}

As can be seen, AEs can be easily altered to achieve different properties in their encoding. The following are some AEs which do not fall into any previous category.

\paragraph{Relational autoencoder}
Basic AEs do not explicitly consider the possible relations among instances. The relational AE \cite{Meng2017} modifies the objective function to take into account the fidelity of the reconstruction of relationships among samples. Instead of just adding a penalty term, the authors propose a weighted sum of the sample reconstruction error and the relation reconstruction error. Notice that this is not the only variation named ``relational autoencoder'' by its authors, different but identically named models are commented in sections \ref{Sec.AEbased} and \ref{Sec.Applications}.

\paragraph{Discriminative autoencoder} Introduced in \cite{DiscriminativeAE}, the discriminative AE uses the class information of instances in order to build a manifold where positive samples are gathered and negative samples are pushed away. As a consequence, this AE performs better reconstruction of positive instances than negative ones. It achieves this by optimizing a different loss function, specifically the hinge loss function used in metric learning. The main objective of this model is object detection.

\subsection{Autoencoder-based architectures for feature learning} \label{Sec.AEbased}

The basic AE can also be used as building block or inspiration for other, more complex architectures dedicated to feature fusion. This section enumerates and briefly introduces the most relevant ones.

\begin{figure}[ht!]
  \centering
  \begin{subfigure}[t]{\columnwidth}
    \centering
    \includegraphics[width=0.6\textwidth]{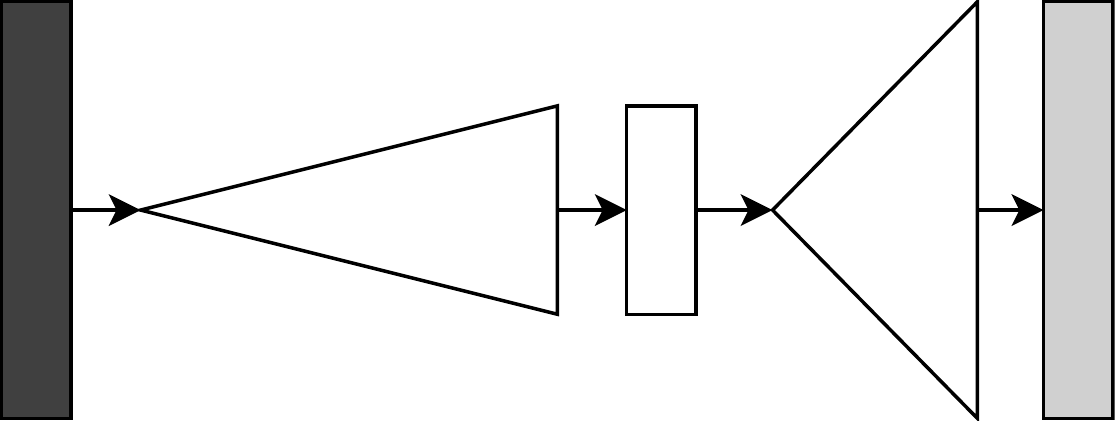}
    \caption{\label{Fig.AETree}AE tree. Triangles represent decision trees.}
  \end{subfigure}

  \vspace{.5em}
  
  \begin{subfigure}[t]{\columnwidth}
    \centering
    \includegraphics[width=0.4\textwidth]{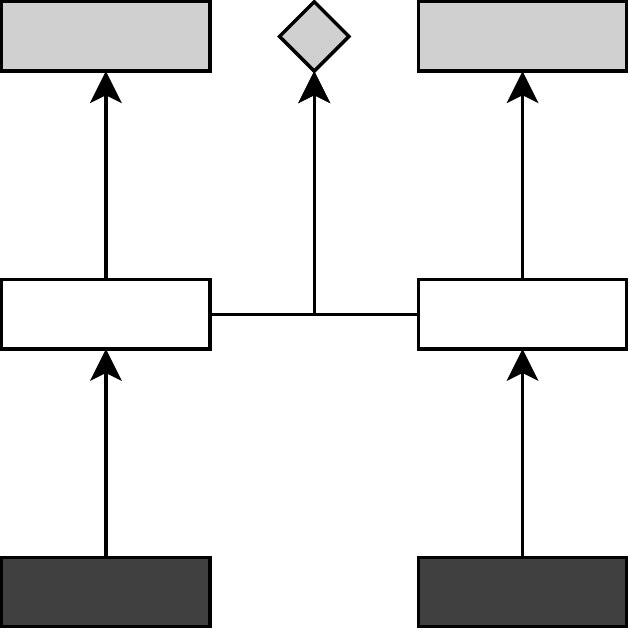}
    \caption{\label{Fig.DualAE}Dual AE. The encodings are coupled by an additional penalty term, represented as a diamond.}
  \end{subfigure}

  \vspace{.5em}
  
  \begin{subfigure}[t]{\columnwidth}
    \centering
    \includegraphics[width=\textwidth]{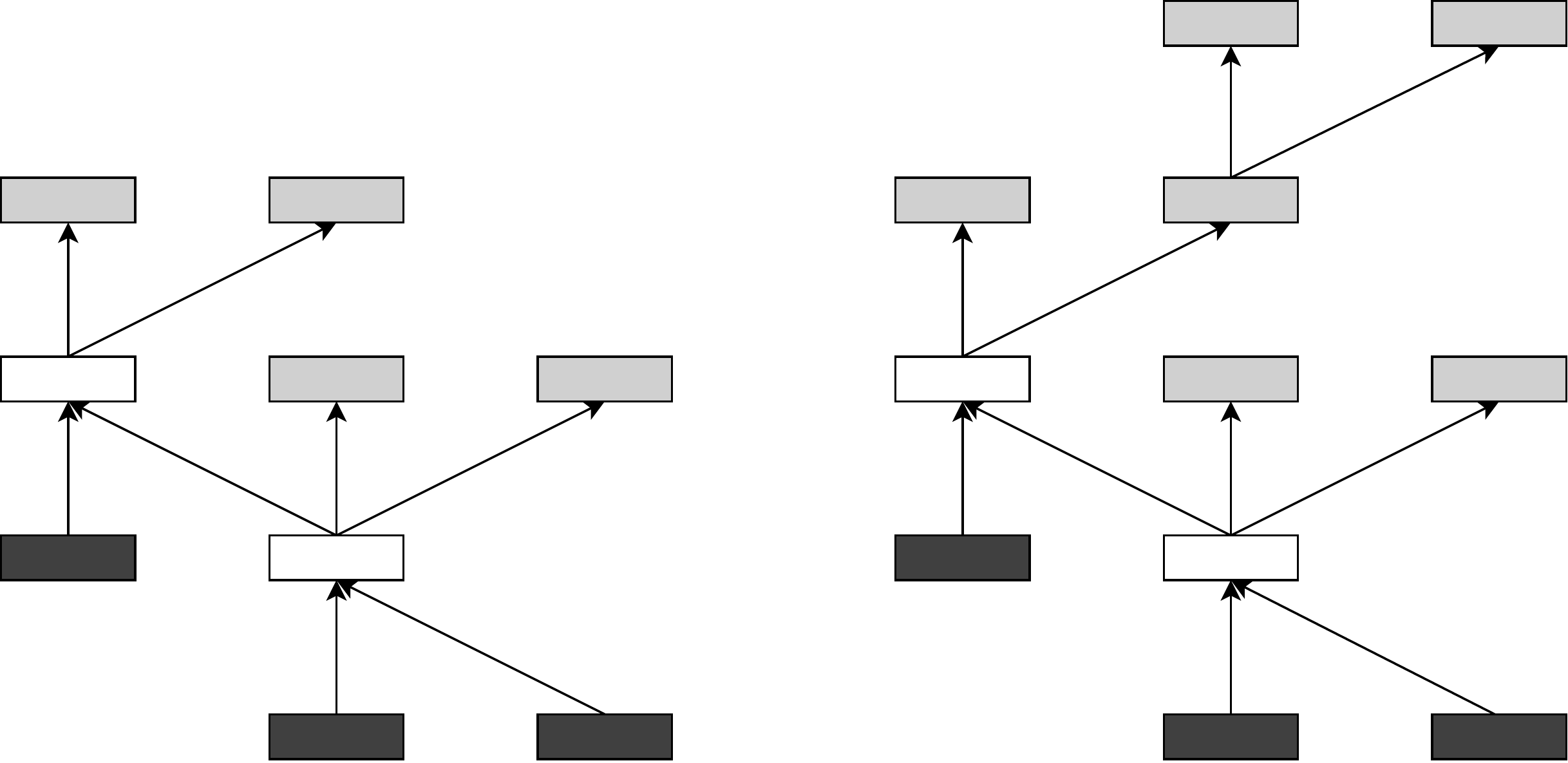}
    \caption{\label{Fig.RecAE}Recursive AE (left), unfolded recursive AE (right)}
  \end{subfigure}
  \caption{\label{Fig.AEarch}Illustrations of autoencoder-based architectures. Each rectangle represents a layer, dark gray fill represents an input, light gray represents output layers and white objects represent hidden layers.}
  
\end{figure}

Autoencoder trees \cite{irsoy_unsupervised_2017} (Fig.~\ref{Fig.AETree}) are encoder-decoder architectures, inspired by neural AEs, where the encoder as well as the decoder are actually decision trees. These trees use soft decision nodes, which means they propagate instances to all their children with different probabilities. 

A dual-autoencoder architecture \cite{He2017} (Fig.~\ref{Fig.DualAE}) attempts to learn two latent representations for problems where variables can be treated as instances and viceversa, e.g. predicting customers' recommendations of items. These two representations are linked by an additional term in the objective function which minimizes their deviation from the training data.

The relational or ``cross-correlation'' AE defined in \cite{MemisevicRelational} incorporates layers where units are combined by multiplication instead of by a weighted sum. This allows it to represent co-ocurrences among components of its inputs.

A recursive AE \cite{RecursiveAE} (Fig.~\ref{Fig.RecAE}) is a tree-like architecture built from AEs, in which new pieces of input are introduced as the model gets deeper. A standard recursive AE attempts to reconstruct only the direct inputs of each encoding layer, whereas an unfolding recursive AE \cite{UnfoldingRecAE} reconstructs all previous inputs from each encoding layer. This architecture is designed to model sentiment in sentences.

\section{Comparison to other feature fusion techniques}\label{Sec.Relationship}

AEs are only several of a very diverse range of feature fusion methods \cite{FeatureFusion}. These can be grouped according to whether they perform supervised or unsupervised learning. In the first case, they are usually known as \textit{distance metric learning} techniques \cite{DistanceMetric}. Some adversarial AEs, as well as AEs preserving class neighborhood structure \cite{NonlinearEmbeddingNN}, can be sorted into this category, since they are able to make use of the class information. However, this section focuses on the latter case, since most AEs are unsupervised and therefore share more similarities with this kind of methods.

A dimensionality reduction technique is said to be \textit{convex} if it optimizes a function which does not have any local optima, and it is \textit{nonconvex} otherwise \cite{DimRecComparative}. Therefore, a different classification of these techniques is into convex and nonconvex approaches. AEs fall into the nonconvex group, since they can attempt to optimize disparate objective functions, and these may present more than one optimum. AEs are also not restrained to the dimensionality reduction domain, since they can produce sparse codes and other meaningful overcomplete representations.

Lastly, feature fusion procedures can be carried out by means of linear or nonlinear transformations. In this section, we aim to summarize the main traits of the most relevant approaches in both of these situations, and compare them to AEs. 

\subsection{Linear approaches}

Principal component analysis is a statistical technique developed geometrically by Pearson \cite{PCA} and algebraically by Hotelling \cite{PCAHotelling}. It consists in the extraction of the \textit{principal components} of a vector of random variables. Principal components are linear combinations of the original variables in a specific order, so that the first one has maximum variance, the second one has maximum possible variance while being uncorrelated to the first (equivalently, orthogonal), the third has maximum possible variance while being uncorrelated to the first and second, and so on. A modern analytical derivation of principal components can be found in \cite{PCABook}.

The use of PCA for dimensionality reduction is very common, and can lead to reasonably good results. It is known that AEs with linear activations that minimize the mean quadratic error learn the principal components of the data \cite{ANNsPCA}. From this perspective, AEs can be regarded as generalizations of PCA. However, as opposed to PCA, AEs can learn nonlinear combinations of the variables and even overcomplete representations of data.

\begin{figure}[hbtp!]
  \begin{enumerate}
  \item \parbox{\columnwidth}{\includegraphics[width=0.9\columnwidth,trim={14em 18em 14em 2em},clip]{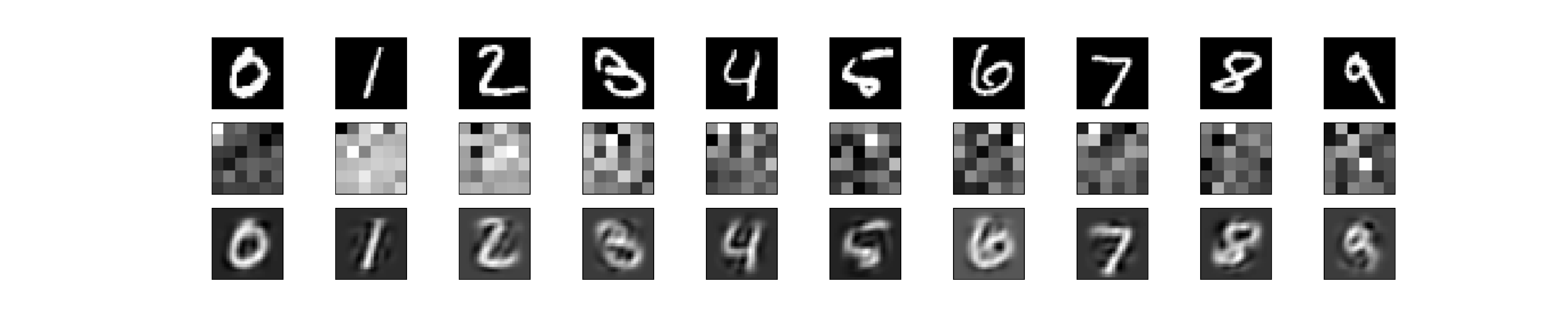}}
  
  \item  \parbox{\columnwidth}{\includegraphics[width=0.9\columnwidth,trim={14em 2em 14em 19em},clip]{pca-36.pdf}}
  
  \item  \parbox{\columnwidth}{\includegraphics[width=0.9\columnwidth,trim={14em 2em 14em 19em},clip]{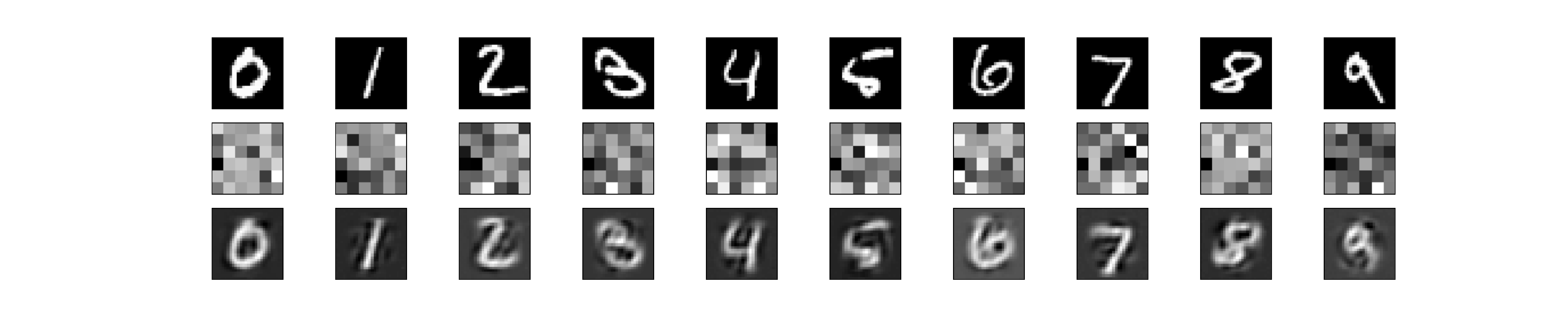}}
  
  \item  \parbox{\columnwidth}{\includegraphics[width=0.9\columnwidth,trim={14em 2em 14em 19em},clip]{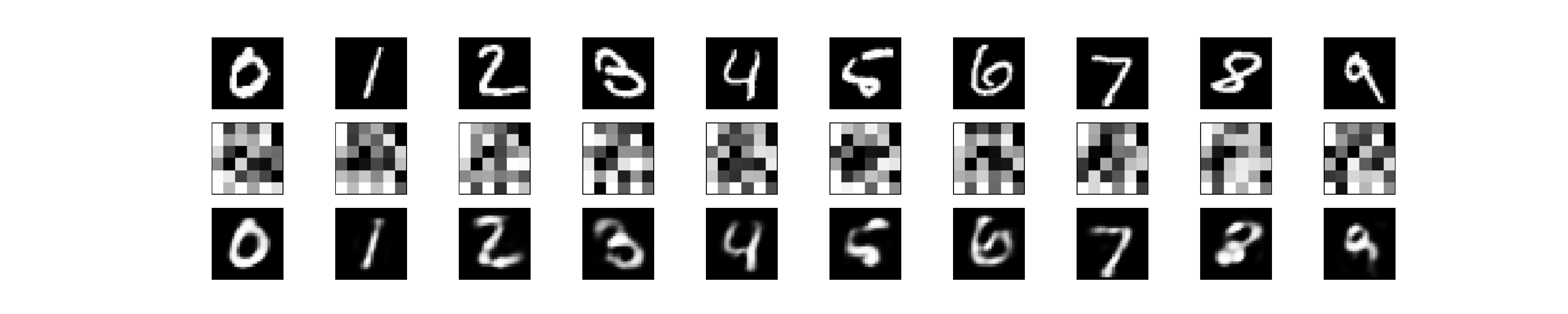}}
  
  \item  \parbox{\columnwidth}{\includegraphics[width=0.9\columnwidth,trim={14em 2em 14em 19em},clip]{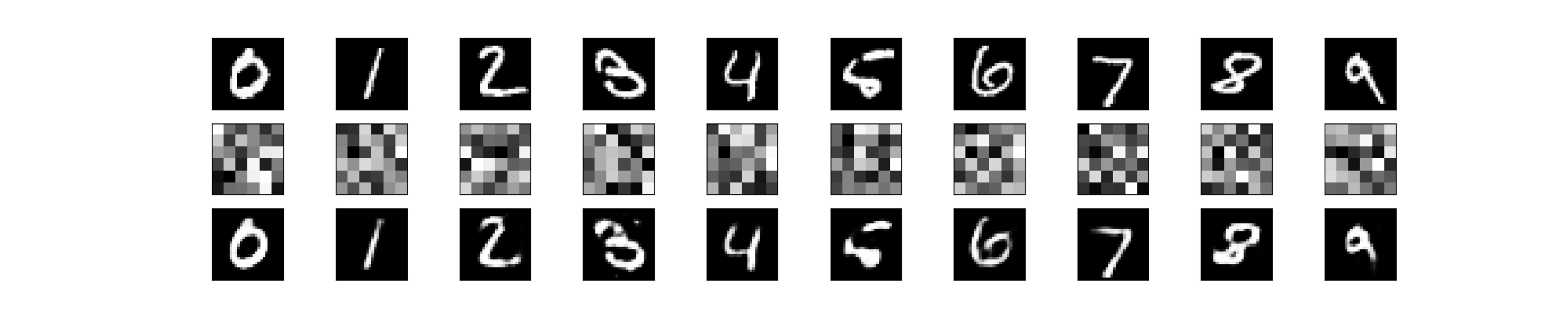}}
  \end{enumerate}
  
  \caption{\label{Fig.AEvsPCA}Row 1 shows test samples, second row corresponds to PCA reconstructions, the third one shows those from a linear AE optimizing MSE, row 4 displays reconstructions from a basic AE with $\tanh$ activation and cross-entropy as loss function, and last row corresponds to a robust AE.}  
\end{figure}

Fig.~\ref{Fig.AEvsPCA} shows a particular occurrence of these facts in the case of the MNIST dataset \cite{MNIST}. Row 1 shows several test samples and the rest display reconstructions built by PCA and some AEs. As can be inferred from rows 2 and 3, linear AEs which optimize MSE learn an approximation of PCA. However, just by adjusting the activation functions and the objective function of the neural network one can obtain superior results (row 4). Improvements over the standard AE such as the robust AE (row 5) also provide higher quality in their reconstructions.

A procedure similar to PCA but from a different theoretical perspective is Factor Analysis (FA) \cite{PCAandFA}, which assumes a set of latent variables or \textit{factors} which are not observable but are linearly combined to produce the observed variables. The difference between PCA and FA is similar to that between the basic AE and the variational AE: the latter assumes that hypothetical, underlying variables exist and cause the observed data. Variational AEs and FA attempt to find the model that best describes these variables, whereas the basic AE and PCA only aim for a lower-dimensional representation.

Linear Discriminant Analysis (LDA) \cite{FisherLDA} is a supervised statistical method to find linear combinations of features that achieve good separation of classes. It makes some assumptions of normality and homoscedasticity over the data, and projects samples onto new coordinates that best discriminate classes. It can be easily seen that AEs are very different in theory to this method: they usually perform unsupervised learning, and they do not necessarily make previous assumptions of the data. In contrast, AEs may not find the best separation of classes but they might encode further meaningful information from the data. Therefore, these techniques may be convenient, each in very different types of problems.

\subsection{Nonlinear approaches}

Kernel PCA \cite{KernelPCA} is an extension of PCA which applies kernel methods in order to extract nonlinear combinations of variables. Since principal components can be computed by projecting samples onto the eigenvectors of the covariance matrix, the kernel trick can be applied in order to calculate the covariance matrix of the data in a higher-dimensional space, given by the kernel function. Therefore, kernel PCA can compute nonlinear combinations of variables and overcomplete representations. The choice of kernel, however, can determine the success of the method and may behave differently with each problem, and hence AEs are a more general and easily applicable framework for nonlinear feature fusion.

Multidimensional Scaling (MDS) \cite{MDS} is a well known technique and a foundation for other algorithms. It consists in finding new coordinates in a lower-dimensional space, while maintaining relative distances among data points as accurately as possible. For this to be achieved, it computes pairwise distances among points and then estimates an origin or zero point for these, which allows to transform relative distances into absolute distances that can be fitted into a real Euclidean space. Sammon mapping \cite{Sammon} modifies the classical cost function of MDS, in an attempt to similarly weigh retaining large distances as well as small ones. It achieves better preservation of local structure than classic MDS, at the cost of giving more importance to very small distances than large ones.  

The approach of MDS to nonlinear feature fusion is opposite to that of AEs, which generally do not directly take into account distances among pairs of samples, and instead optimize a global measure of fitness. However, the objective function of an AE can be combined with that of MDS in order to produce a nonlinear embedding which considers pairwise distances among points \cite{EmbeddingAEReg}.

Isomap \cite{Isomap} is a manifold learning method which extends MDS in order to find coordinates that describe the actual degrees of freedom of the data while preserving distances among neighbors and geodesic distances between the rest of points. In addition, Locally Linear Embedding (LLE) \cite{LLE} has a similar goal, to learn a manifold which preserves neighbors, but a very different approach: it linearly reconstructs each point from its neighbors in order to maintain the local structure of the manifold.

Both of these techniques can be compared to the contractive AE, as it also attempts to preserve the local behavior of the data in its encoding. Denoising AEs may also be indirectly forced to learn manifolds, when they exist, and corrupted examples will be projected back onto their surface \cite{vincent2010stacked}. However, AEs are able to map new instances onto the latent space after they have been trained, a task Isomap and LLE are not designed for.

Laplacian Eigenmaps \cite{LaplacianEigenmaps} is a framework aiming to retain local properties as well. It consists in constructing an adjacency graph where instances are nodes and neighbors are connected by edges. Then, a weight matrix similar to an adjacency matrix is built. Last, eigenvalues and eigenvectors are obtained for the Laplacian matrix associated to the weight matrix, and those eigenvectors (except 0) are used to compute new coordinates for each point. As previously mentioned, AEs do not usually consider the local structure of the data, except for contractive AEs and further regularizations which incorporate measures of local properties into the objective function, such as Laplacian AEs \cite{LaplacianAE}.

A Restricted Boltzmann Machine (RBM) \cite{DLBookRBM}, introduced originally as \textit{harmonium} in \cite{Harmonium}, is an undirected graphical model, with one visible layer and one hidden layer. They are defined by a joint probability distribution determined by an energy function.
However, computing probabilities is unfeasible since the distribution is intractable, and they have been proved to be hard to simulate \cite{RBMHard}. Instead, Contrastive Divergence \cite{ContrastiveDivergence} is used to train an RBM. RBMs are an alternative to AEs for greedy layer-wise initialization of weights in ANNs including AEs. AEs, however, are trained with more classical methods and are more easily adaptable to different tasks than RBMs.

\section{Applications in feature learning and beyond}\label{Sec.Applications}

The ability of AEs to perform feature fusion is useful for easing the learning of predictive models, improving classification and regression results, and also for facilitating unsupervised tasks that are harder to conduct in high-dimensional spaces, such as clustering. Some specific cases of these applications are portrayed within the following subsections, including:
\begin{itemize}
  \setlength\itemsep{-.2em}
\item Classification: reducing or transforming the training data in order to achieve better performance in a classifier.
\item Data compression: training AEs for specific types of data to learn efficient compressions.
\item Detection of abnormal patterns: identification of discordant instances by analyzing generated encodings.
\item Hashing: summarizing input data onto a binary vector for faster search.
\item Visualization: projecting data onto 2 or 3 dimensions with an AE for graphical representation.
\item Other purposes: further applications of AEs.
\end{itemize}

\subsection{Classification}
Using any of the AE models described in Section \ref{Sect.AEforFF} to improve the output of a classifier is something very common nowadays. Here only a few but very representative case studies are referenced.

Classifying tissue images to detect cancer nuclei is a very complicated accomplishment, due to the large size of high-resolution pathological images and the high variance of the fundamental traits of these nuclei, e.g. its shape, size, etc. The authors of \cite{Xu2016} introduce a method, based on stacked DAEs to produce higher level and more compact features, which eases this task.

Multimodal/Multiview learning \cite{MultiviewLearning} is a rising technique which also found considerable support in AEs. The authors of \cite{tian_learning_2016} present a general procedure named \textit{Orthogonal Autoencoder for Multi-View}. It is founded on DAEs to extract private and shared latent feature spaces, with an added orthogonality constraint to remove unnecessary connections. In \cite{liu_multimodal_2016} the authors propose the MSCAE (\textit{Multimodal Stacked Contractive Autoencoder}), an application-specific model fed with text, audio and image data to perform multimodal video classification. 

Multilabel classification \cite{Charte:SB-MLC} (MLC) is another growing machine learning field. MLC algorithms have to predict several outputs (labels) linked to each input pattern. These are usually defined by a high-dimensional feature vector and a set of labels as output which tend to be quite large as well. In \cite{MLC-DL} the authors propose an AE-based method named C2AE (\textit{Canonical Correlated AutoEncoder}), aimed to learn a compressed feature space while establishing relationships between the input and output spaces.

Text classification following a semi-supervised approach by means of AEs is introduced in \cite{Xu2017VariationalAF}. A model called SSVAE (\textit{Semi-supervised Sequential Variational Autoencoder}) is presented, mixing a Seq2Seq \cite{Seq2Seq} structure and a sequential classifier. The authors state that their method outperforms fully supervised methods.

Classifiers based on AEs can be grouped in ensembles in order to gain expressive power, but some diversity needs to be introduced. Several means of doing so, as well as a proposal for Stacked Denoising Autoencoding (SDAE) classifiers can be found in \cite{Alvear_Sandoval_2018}. This method has set a new performance record in MNIST classification.

\subsection{Data compression}
Since AEs are able to reconstruct the inputs given to them, an obvious application would be compressing large amounts of data. However, as we already know, the AE output is not perfect, but an approximate reconstruction of the input. Therefore, it is useful only when lossy compression is permissible. 

This is the usual scenario while working with images, hence the popularity of the JPEG \cite{JPEG} graphic file format. It is therefore not surprising that AEs have been successfully applied to this task. This is the case of \cite{tan_using_2011}, where the performance of several AE models compressing mammogram image patches is analyzed. A less specific goal can be found in \cite{AEImageCompress}. It proposes a model of AE named SWTA AE (\textit{Stochastic Winner-Take-All Auto-Encoder}), a variation of the sparse AE model, aimed to work as a general method able to achieve a variable ratio of image compression.

Although images could be the most popular data compressed by means of AEs, these have also demonstrated their capacity to work with other types of information as well. For instance:
\begin{itemize}
    \item In  \cite{AEBioCompress} the authors suggest the use of AEs to compress biometric data, such as blood pressure or heart rate, retrieved by wearable devices. This way  battery life can be extended while time transmission of data is reduced. 
    
    \item Language compression is the goal of ASC (\textit{Autoencoding Sentence Compression}), a model introduced in \cite{VAESentenceCompress}. It is founded on a variational AE, used to draw sentences from a language modeled with a certain distribution.
    
    \item High-resolution time series of data, such as measurements taken from service grids (electricity, water, gas, etc.), tend to need a lot of space. In \cite{RAETSCompress} the APRA (\textit{Adaptive Pairwise Recurrent Encoder}) model is presented, combining an AE and a LSTM to successfully compress this kind of information.
\end{itemize}

Lossy compression is assumed to be tolerable in all these scenarios, so the approximate reconstruction process of the AE does not hinder the main objective in each case.

\subsection{Detection of abnormal patterns}
Abnormal patterns are samples present in the dataset that clearly differ from the remaining ones. The distinction between \textit{anomalies} and \textit{outliers} is usually found in the literature, although according to Aggarwal \cite{OutlierAnalysis} these terms, along with \textit{deviants}, \textit{discordants} or \textit{abnormalities}, refer to the same concept.

The telemetry obtained from spacecrafts is quite complex, made up of hundreds of variables. The authors of \cite{sakurada_anomaly_2014} propose the use of basic and denoising AEs for facing anomaly detection taking advantage of the nonlinear dimensionality reduction ability of these models. The comparison with both PCA and Kernel PCA demonstrates the superiority of AEs in this task.

The technique introduced in \cite{Chen2017OutlierDW} aims to improve the detection of outliers. To do so, the authors propose to create ensembles of AEs with random connections instead of fully connected layers. Their model, named RandNet (\textit{Randomized Neural Network for Outlier Detection}), is compared against four classic outlier detection methods achieving an outstanding performance.

A practical application of abnormal pattern detection with AEs is the one proposed in \cite{Castellini2017}. The authors of this work used a DAE, trained with a benchmark dataset, to identify fake twitter accounts. This way legitimate followers can be separated of those that are not.

\subsection{Hashing}\label{Sect.Hashing}
Hashing \cite{HashingReview2017} is a very common technique in computing, mainly to create data structures able to offer constant access time to any element (\textit{hash tables}) and to provide certain guarantees in cryptography (\textit{hash values}). A special family of hash functions are those known as \textit{Locality Sensitive Hashing} (LSH) \cite{LSH}. They have the ability to map data patterns to lower dimensional spaces while maintaining some topological traits, such as the relative distance between these patterns. This technique is very useful for some applications, such as similar document retrieval. AEs can be also applied in these same fields.

Salakhutdinov and Hinton demonstrated in \cite{SemanticHashing} how to perform what they call \textit{semantic hashing} through a multi-layer AE. The fundamental idea is to restrict the values of the encoding layer units so that they are binary. In the example proposed in this study that layer has 128 or 20 units, sequences of ones and zeroes that are interpreted as an address. The aim is to facilitate the retrieval of documents, as noted above. The authors show how this technique offers better performance than the classic TF-IDF \cite{TFIDF} or LSH.

Although the approach to generate the binary AE is different from the previous one, since they achieve hashing with binary AEs helped by MAC (\textit{Method of Auxiliary Coordinates}) \cite{MethodAuxiliaryCoordinates}, the proposal in \cite{carreira-perpinan_hashing_2015} is quite similar. The encoding layer produces a string of zeroes and ones, used in this case to conduct fast search of similar images in databases.

\subsection{Data visualization}\label{Sec.Visualization}

Understanding the nature of a given dataset can be a complex task when it posesses many dimensions. Data visualization techniques \cite{InfoVis} can help analyze the structure of the data. One way of visualizing all instances in a dataset is to project it onto a lower-dimensional space which can be represented graphically.

\begin{figure}[h!]
  \centering
  \includegraphics[width=.95\columnwidth]{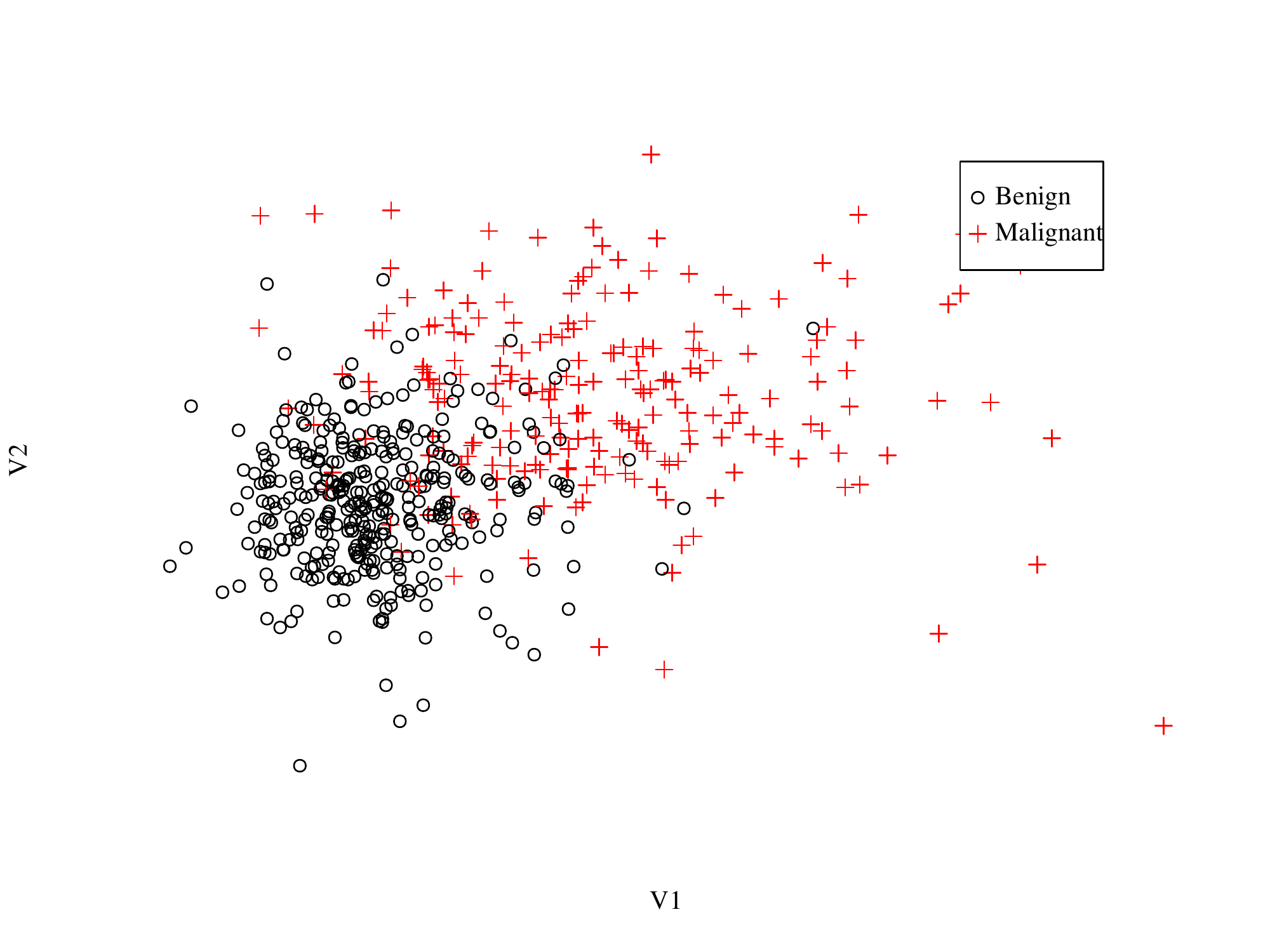}
  \caption{\label{Fig.Cancer}Example visualization of the codifications of a Cancer dataset generated with a basic AE with weight decay.}
\end{figure}

A particular useful case of AEs are those with a 2 or 3-variable encoding \cite{hinton_reducing_2006}. This allows the generated codifications of samples to be displayed in a graphical representation such as the one in Fig.~\ref{Fig.Cancer}.

The original data \cite{WDBC} has 30 variables describing each pattern. Each data point is linked to one of two potential cancer diagnosis (classes), Benign and Malignant. These have been used in Fig.~\ref{Fig.Cancer} to better show the separation between the two classes, but the V1 and V2 variables have been produced by the AE in an unsupervised fashion. Different projections could be obtained by adjusting the AE parameters.

\subsection{Other applications of autoencoders}
Beyond the specific applications within the four previous categories, which can be considered as usual in terms of the use of AEs, these find to be useful in many other cases. The following are just a few specific examples.

Holographic images \cite{DigitalHolography} are a useful resource to store information in a fast way. However, retrieval of data has to face a common obstacle as is image degradation by the presence of speckle noise. In \cite{Shimobaba2017} an AE is trained with original holographic images as well as with degraded images, aiming to have a decoder able to reconstruct deteriorated examples. The denoising of images is also the goal of the method introduced in \cite{DenoisingMedicalImages}, although in this case they are medical images and the AE method is founded on convolutional denosing AEs.

The use of AEs to improve automatic speech recognition (ASR) systems has been also studied in late years. The authors of \cite{SpeechDAE} rely on a DAE to reduce the noise and thus perform speech recognition enhancement. Essentially, the method gives the deep DAE noisy speech samples as inputs while the reference outputs are clean. A similar procedure is followed in \cite{ReverberantSpeechDAE}, although in this case the problem present in the speech samples is reverberation. ASR is specially challenging when faced with whispered speech, as described in \cite{WhisperedSppechDAE}. Once more, a deep DAE is the tool to improve results from classical approaches.

The procedure to curate biological databases is very expensive, so usually machine learning methods such as SVD (\textit{Singular Value Decomposition}) \cite{SVD} are applied to help in the process. In \cite{ANNsPCA2} this classical approach is compared with the use of deep AEs, reaching as conclusion that the latter is able to improve the results.

The authors of  \cite{Hong2015} aim to perform multimodal fusion by means of deep AEs, specifically proposing a Multimodal Deep Autoencoder (MDA). The goal is to perform human pose recovery from video \cite{HumanPoseRecovery}. To do so, two separate AEs are used to obtain high-level representations of 2D images and 3D human poses. Connecting these two AEs, a two-layer ANN carries out the mapping between the two representations.

Tagging digital resources, such as movies and products \cite{TagRecommendation} or even questions in forums \cite{QUINTA}, helps the users in finding the information they are interested in, hence the importance in designing tag recommendation systems. The foundation of the approach in \cite{RelationalDAE} is an AE variation named  RSDAE (\textit{Relational Stacked Denoising Autoencoder}). This AE works as a graphical model, combining the learning of high-level features with relationships among items.

AEs are also scalable to diverse applications with big data, where the stacking of networks acquires notable importance \cite{SurveyDLbigdata}. Multi-modal AEs and Tensor AEs are some examples of variants developed in this field.

\section{Guidelines, software and examples on autoencoder design}\label{Sec.HowToChoose}

This section attempts to guide the user along the process of designing an AE for a given problem, reviewing the range of choices the user has and their utility, then summarizing the available software for deep learning and outlining the steps needed to implement an AE. It also provides a case study with the MNIST dataset where the impact of several parameters of AEs is explored, as well as different AE types with identical parameter settings.

\subsection{Guidelines}

\begin{figure}[ht!]
  \centering
  \resizebox {\columnwidth} {!} {
  \begin{tikzpicture}[
      font=\scriptsize,
      every node/.style = {shape=rectangle,
        draw, align=center},
      level 1/.style={sibling distance=20ex},
      level 2/.style={grow=down, anchor=west, draw=gray, xshift=-1em,
        edge from parent path={([xshift=.5em]\tikzparentnode.south west) |- (\tikzchildnode.west)}},
      level 3/.style={grow=down, anchor=west, draw=gray, xshift=-1em,
        edge from parent path={([xshift=.5em]\tikzparentnode.south west) |- (\tikzchildnode.west)}},
      level distance=6ex,
      first/.style={level distance=4ex,xshift=-1.1em},
      second/.style={level distance=8ex,xshift=-1.1em},
      third/.style={level distance=12ex,xshift=-1.1em},
      fourth/.style={level distance=16ex,xshift=-1.1em},
      fifth/.style={level distance=20ex,xshift=-1.1em},
      sixth/.style={level distance=24ex,xshift=-1.1em}
    ],
      
      \node {Choices when designing AEs}
      child { node {Architecture}
        child[first]  { node {No. of layers} }
        child[second] { node {No. of units} }
        child[third]  { node {Unit type} 
          child[first]  { node {Fully connected} }
          child[second] { node {Convolutional} }
          child[third]  { node {LSTM} }
        }
      }
      child { node {Loss function}
        child[first]  { node {Main term}
          child[first]  { node {MSE} }
          child[second] { node {Cross-entropy} }
          child[third]  { node {Correntropy} }
        }
        child[fifth] { node {Regularizations}
          child[first]  { node {Sparsity} }
          child[second] { node {Contraction} }
          child[third]  { node {Weight decay} }
        }
      }
      child { node {Activations}
        child[first]  { node {Tanh} }
        child[second] { node {Sigmoid} }
        child[third]  { node {ReLU} }
        child[fourth] { node {SELU} }
        child[fifth]  { node {Linear} }
        child[sixth]  { node {...} }
      };
  \end{tikzpicture}
  }
  \caption{\label{Fig.AEchoices}Summary of choices when designing an AE}

\end{figure}
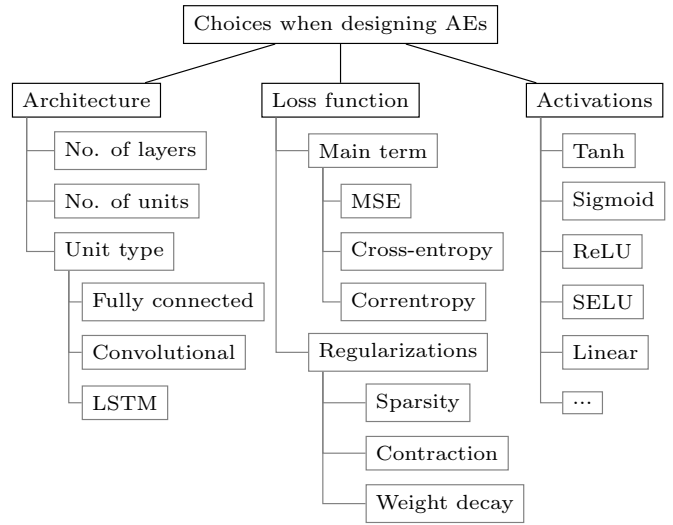

When building an AE for a specific task, it is convenient to take into consideration the modifications studied in Section~\ref{Sect.AEforFF}. There is no need to choose just one of those, most of them can actually be combined in the same AE. For instance, one could have a stacked denoising AE with weight decay and sparsity regularizations. A schematic summary of these can be viewed in Fig.~\ref{Fig.AEchoices}.

\paragraph{Architecture} Firstly, one must define the structure of the AE, especially the length of the encoding layer. This is a fundamental step that will determine whether the training process can lead it to a good codification. If the length of the encoding is proportionally very low with respect to the number of original variables, training a deep stacked AE should be considered. In addition, convolutional layers are generally better performant with image data, whereas LSTM encoders and decoders would be preferable when modeling sequences. Otherwise, fully connected layers should be chosen. 

\paragraph{Activations and loss function} Activation functions that will be applied within each layer have to be decided according to the loss function which will be optimized. For example, a sigmoid-like function such as the logistic or $\tanh$ is generally a reasonable choice for the encoding layer, the latter being usually preferred due to its greater gradients. This does not need to coincide with the activation in the output layer. Placing a linear activation or ReLU at the output can be sensible when using mean squared error as reconstruction error, while a logistic activation would be better combined with the cross-entropy error and normalized data, since it outputs values between 0 and 1.

\paragraph{Regularizations} On top of that, diverse regularizations may be applied that will lead the AE to improve its encoding following certain criteria. It is generally advisable to add a small weight decay in order to prevent it from overfitting the training data. A sparse codification is useful in many cases and adds more flexibility to the choice of structure. Additionally, a contraction regularization may be valuable if the data forms a lower-dimensional manifold. \\

As seen in previous sections, AEs provide high flexibility and can be further modified for very different applications. In the case that the standard components do not fit the desired behavior, one must study which of those can be replaced and how, in order to achieve it.

\subsection{Software}

There exists a large spectrum of cross-platform, open source implementations of deep learning methods which allow for the construction and training of AEs. This section summarizes the most popular frameworks, enumerates some specific implementations of AEs, and provides an example of use where an AE is implemented on top of one of these frameworks.

\subsubsection{Available frameworks and packages}

\paragraph{Tensorflow \cite{Tensorflow}} Developed by Google, Tensorflow has been the most influential deep learning framework. It is based on the concept of \textit{data flow graphs}, where nodes represent mathematical operations and multidimensional data arrays travel through the edges. Its core is written in C++ and interfaces mainly with Python, although there are APIs for Java, C and Go as well. 

\paragraph{Caffe \cite{Caffe}} Originating at UC Berkeley, Caffe is built in C++ with speed and modularity in mind. Models are defined in a descriptive language instead of a common programming language, and trained in a C++ program.

\paragraph{Torch \cite{torch}} It is a Lua library which promises speed and flexibility, but the most notorious feature is its large ecosystem of community-contributed tutorials and packages.

\paragraph{MXNet \cite{MXNet}} This project is currently held at the Apache Incubator for incoming projects into the Apache Foundation. It is written in C++ and Python, and offers APIs in several additional languages, such as R, Scala, Perl and Julia. MXNet provides flexibility in the definition of models, which can be programmed symbolically as well as imperatively.

\paragraph{Keras \cite{Keras}} Keras is a higher-level library for deep learning in Python, and can rely on Tensorflow, Theano, MXNet or Cognitive Toolkit for the underlying operations. It simplifies the creation of deep learning architectures by providing several shortcuts and predefined utilities, as well as a common interface for several deep learning toolkits.\\

In addition to the previous ones, other well known deep learning frameworks are Theano~\cite{Theano}, Microsoft Cognitive Toolkit (CNTK\footnote{\url{https://docs.microsoft.com/cognitive-toolkit/}}) and Chainer~\footnote{\url{https://chainer.org/}}.

Setting various differences apart, all of these frameworks present some common traits when building AEs. Essentially, the user has to define the model layer by layer, placing activations where desired. When establishing the objective function, they will surely include the most usual ones, but uncommon loss functions such as correntropy or some regularizations such as contraction may need to be implemented additionally.

Very few pieces of software have specialized in the construction of AEs. Among them, there is an implementation of the sparse AE available in packages Autoencoder \cite{CRANautoencoder} and SAENET \cite{SAENET} of the CRAN repository for R, as well as an option for easily building basic AEs in H2O\footnote{\url{http://docs.h2o.ai}}. The yadlt\footnote{\url{https://deep-learning-tensorflow.readthedocs.io/}} library for Python implements denoising AEs and several ways of stacking AEs.

\subsubsection{Example of use}

For the purposes of the case study in Section~\ref{Sec.CaseStudy}, some simple implementations of different shallow AEs have been developed and published on a public code repository under a free software license\footnote{\url{https://github.com/fdavidcl/ae-review-resources}}. In order to use these scripts, the machine will need to have Keras and Tensorflow installed. This can be achieved from a Python package manager, such as \texttt{pip} or \texttt{pipenv}, or even general package managers from some Linux distributions.

In the provided repository, the reader can find four scripts dedicated to AEs and one to PCA. Among the first ones, \texttt{autoencoder.py} defines the Keras model for a given AE type with the specified activation for the encoding layer. For its part, \texttt{utils.py} implements regularizations and modifications in order to be able to define basic, sparse, contractive, denoising and robust AEs.

Executable scripts are \texttt{mnist.py} and \texttt{cancer.py}. The first trains any AE with the MNIST dataset and outputs a graphical representation of the encoding and reconstruction of some test instances, whereas the latter needs the Wisconsin Breast Cancer Diagnosis (WDBC) dataset in order to train an AE for it. To use them, just call the Python interpreter with the script as an argument, e.g. \texttt{python mnist.py}.

In order to modify the learned model in one of these scripts, the user will need to adjust parameters in the construction of an \texttt{Autoencoder} object. The following is an example which will define a sparse denoising AE:
\begin{verbatim}
dsae = Autoencoder(
  input_dim    = 784,   encoding_dim = 36,
  weight_decay = False, sparse       = True,
  contractive  = False, denoising    = True,
  robust       = False, activation   = "tanh"
)
\end{verbatim}
Other numerical parameters for each AE type can be further customized inside the \texttt{build} method. The training process of this AE can be launched via a \texttt{MNISTTrainer} object:
\begin{verbatim}
MNISTTrainer(dsae).train(
  optimizer = "adam", epochs = 50,
  loss = losses.binary_crossentropy
).predict_test()
\end{verbatim}
Finally, running the modified script will train the AE and output some graphical representations.

The \texttt{Autoencoder} class can be reused to train AEs with other datasets. For this, one would need to implement funtionality analogous to the \texttt{MNISTTrainer} class, which loads and prepares data, which is provided to the AE model to be trained. A different example can be found in the \texttt{CancerTrainer} class for the WDBC dataset.

\subsection{Case study: handwritten digits}\label{Sec.CaseStudy}
In order to offer some insight into the behavior of the main kinds of AE that can be applied to the same problem, as well as some of the key points in their configuration, we can study the resulting codifications and reconstructions when training them with the well known dataset of handwritten digits MNIST \cite{MNIST}. To do so, we have trained several AEs with the 60~000 training instances, and have obtained reconstructions for the first test instance of each class. Input values, originally ranging from 0 to 255, have been scaled to the $[0,1]$ interval.

By default, the architecture of every AE has been as follows: a 784-unit input layer, a 36-unit encoding layer with $\tanh$ activation and a 784-unit output layer with sigmoid activation. They have been trained with the RMSProp algorithm for a total of 60 epochs and use binary cross-entropy as their reconstruction error, except for the robust AE which uses its own loss function, correntropy. They are all provided identical weight initializations  and hyperparameters.

Firstly, the performance impact of the encoding length and the optimizer  is studied. Next, changes in the behavior of a standard AE due to different activation functions are analyzed. Lastly, the main AE models for feature fusion are compared sharing a common configuration. Scripts that were used to generate these results were implemented in Python, with the Keras library over the Tensorflow backend.

\subsubsection{Settings of encoding length}
\begin{figure}[htbp]
  \centering
  \includegraphics[width=\columnwidth,trim={16em 18em 14em 2em},clip]{basic-36-tanh-rmsprop-xent.pdf}
  \includegraphics[width=\columnwidth,trim={16em 2em 14em 19em},clip]{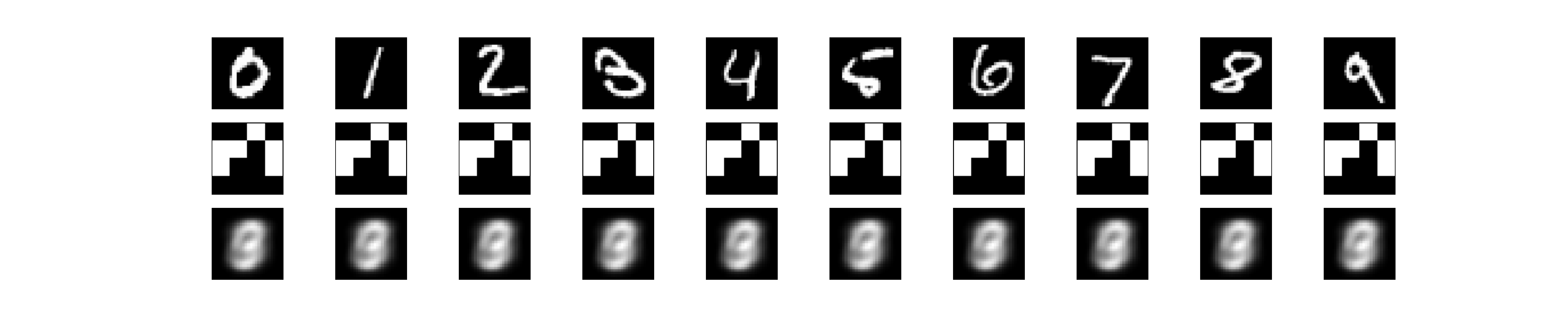}
  \includegraphics[width=\columnwidth,trim={16em 2em 14em 19em},clip]{basic-36-tanh-rmsprop-xent.pdf}
  \includegraphics[width=\columnwidth,trim={16em 2em 14em 19em},clip]{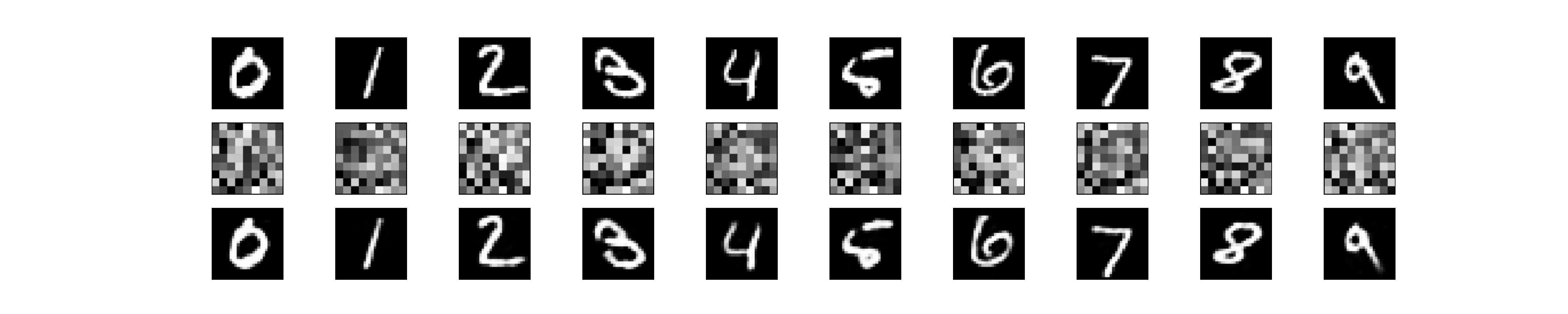}
  \includegraphics[width=\columnwidth,trim={16em 2em 14em 19em},clip]{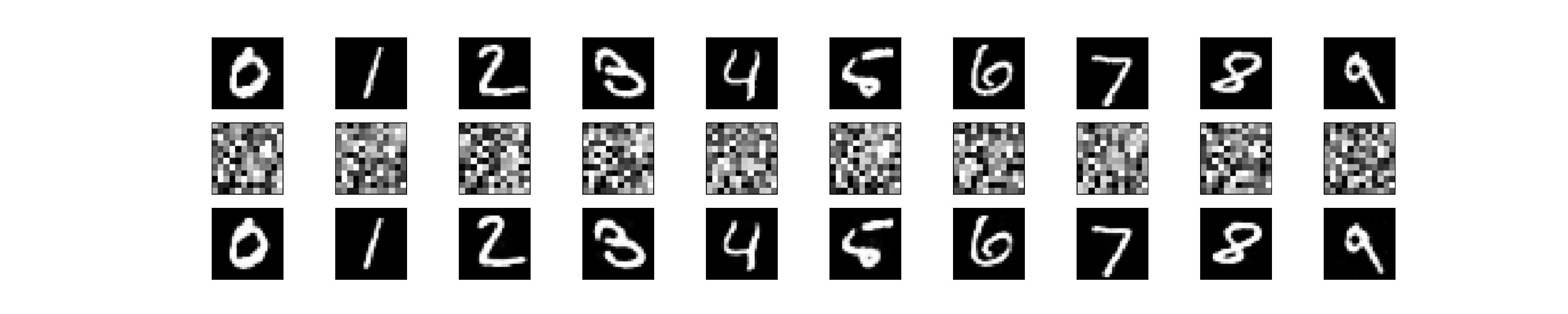}
  \caption{\label{Fig.enclength}First row: test samples; Remaining rows: reconstructions obtained with 16, 36, 81 and 144 units in the encoding layer, respectively.}
\end{figure}

As discussed previously, the number of units in the encoding layer can determine whether the AE is able to learn a useful representation. This fact is captured in Fig.~\ref{Fig.enclength}, where an encoding of 16 variables is too small for the shallow AE to be successfully trained with the default configuration, but a 36-variable codification achieves reasonably good reconstructions. The accuracy of these can be improved at the cost of enlarging the encodings, as can be seen with the 81 and 144-variable encodings. Square numbers were chosen for the encoding lengths for easier graphical representation, as will be seen in Section~\ref{Sec.case.models}, but any other length would have been as valid.

\subsubsection{Comparison of optimizers}

As introduced in Section~\ref{Sec.Training}, AEs can use several optimization methods, usually based on SGD. Each variant attempts to improve SGD in a different way, habitually by accumulating previous gradients in some way or dynamically adapting parameters such as the learning rate. Therefore, they will mainly differ in their ability to converge and their speed in doing so.

\begin{figure}[htbp]
  \includegraphics[width=\columnwidth,trim={0 0 0 6em},clip]{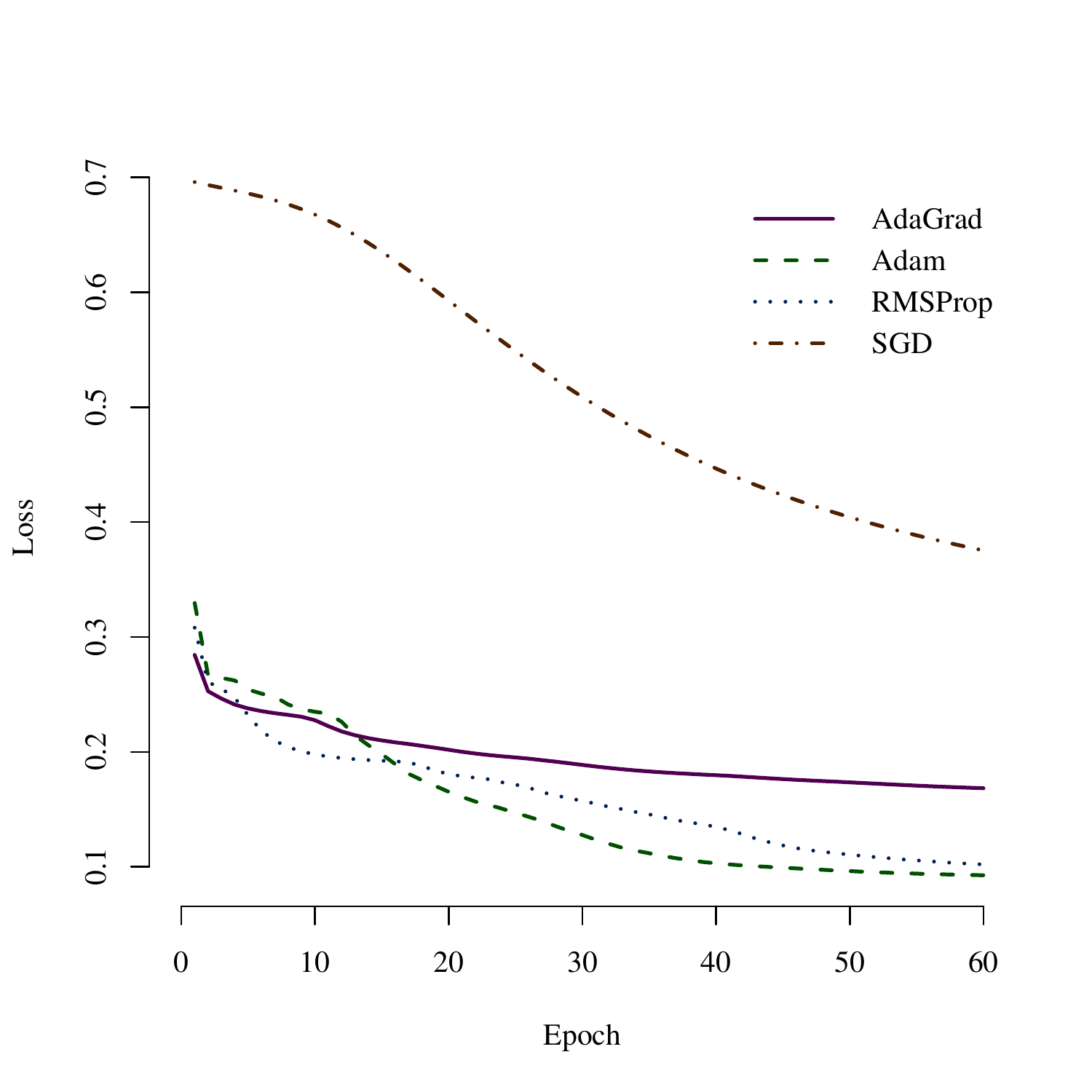}
  \caption{\label{Fig.opt.loss}Evolution of the loss function when using several optimizers.}
\end{figure}

The optimizers used in these examples were baseline SGD, AdaGrad, Adam and RMSProp. Their progressive improvement of the objective function through the training phase is compared in Fig.~\ref{Fig.opt.loss}. It is easily observed that SGD variants vastly improve the basic method, and Adam obtains the best results among them, being closely followed by AdaGrad. The speed of convergence seems slightly higher in Adam as well.

\begin{figure*}[htbp]
  \centering
  \begin{subfigure}{0.48\textwidth}
    \includegraphics[width=\textwidth,trim={14em 18em 14em 2em},clip]{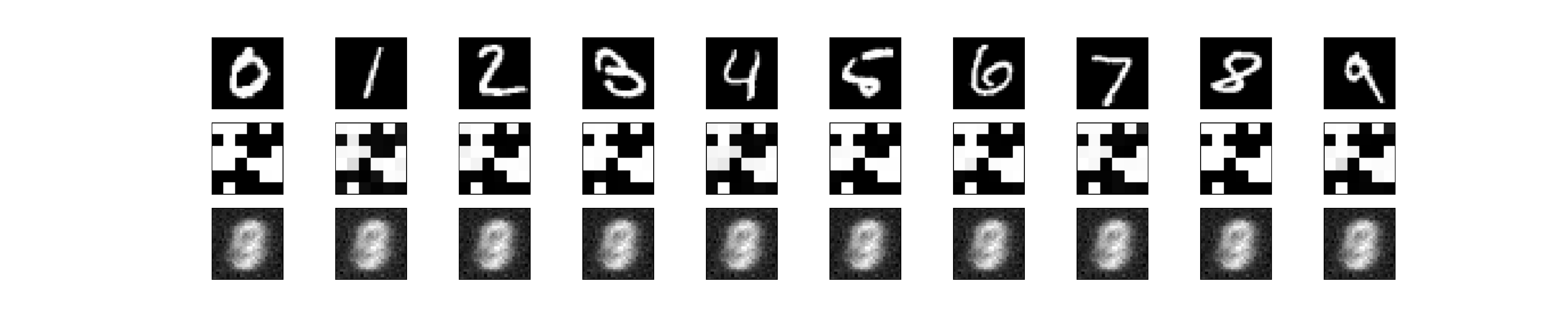}
    \caption{\label{Fig.opt.orig}Test inputs}
  \end{subfigure}

  \vspace{1em}
  
  \begin{subfigure}{0.48\textwidth}
    \includegraphics[width=\textwidth,trim={14em 2em 14em 18em},clip]{basic-36-tanh-sgd-xent.pdf}
    \caption{\label{Fig.opt.sgd}SGD}
  \end{subfigure}
  \begin{subfigure}{0.48\textwidth}
    \includegraphics[width=\textwidth,trim={14em 2em 14em 19em},clip]{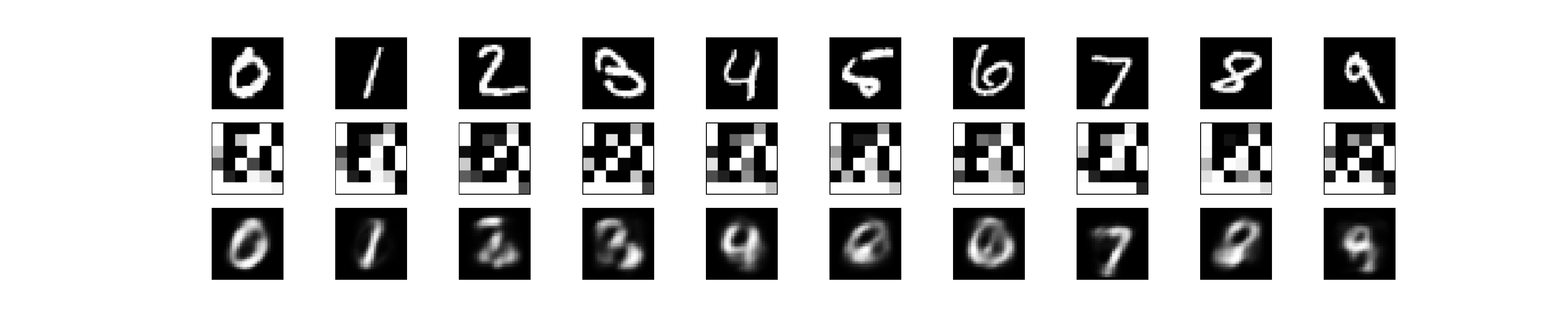}
    \caption{\label{Fig.opt.adagrad}AdaGrad}
  \end{subfigure}

  \vspace{1em}
  
  \begin{subfigure}{0.48\textwidth}
    \includegraphics[width=\textwidth,trim={14em 2em 14em 19em},clip]{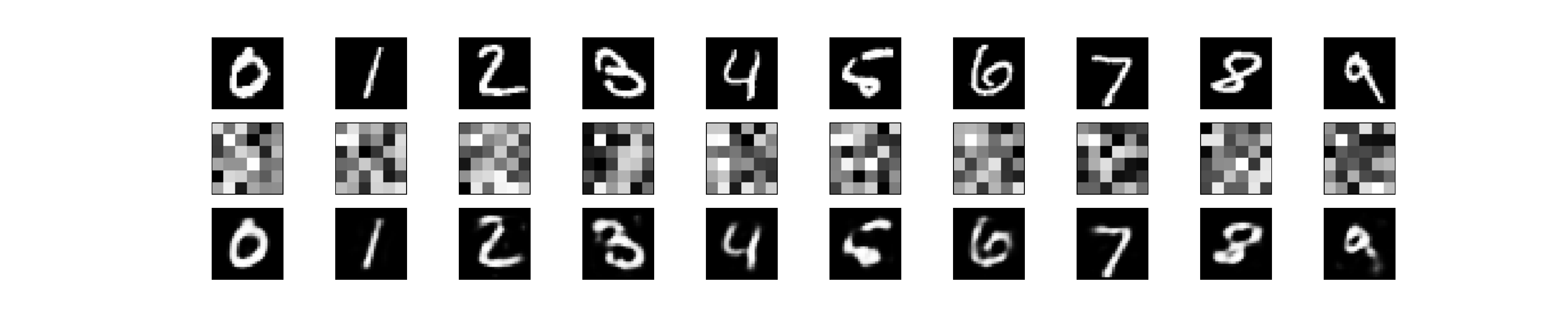}
    \caption{\label{Fig.opt.adam}Adam}
  \end{subfigure}
  \begin{subfigure}{0.48\textwidth}
    \includegraphics[width=\textwidth,trim={14em 2em 14em 19em},clip]{basic-36-tanh-rmsprop-xent.pdf}
    \caption{\label{Fig.opt.rmsprop}RMSProp}
  \end{subfigure}
 
  \caption{\label{Fig.optimizers}Test samples and reconstructions obtained with different optimizers.}
\end{figure*}

\begin{figure*}[htbp]
  \centering
  \begin{subfigure}{0.48\textwidth}
    \includegraphics[width=\textwidth,trim={14em 18em 14em 2em},clip]{basic-36-tanh-sgd-xent.pdf}
    \caption{\label{Fig.act.orig}Test inputs}
  \end{subfigure}

  \vspace{1em}
  
  \begin{subfigure}{0.48\textwidth}
    \includegraphics[width=\textwidth,trim={14em 2em 14em 19em},clip]{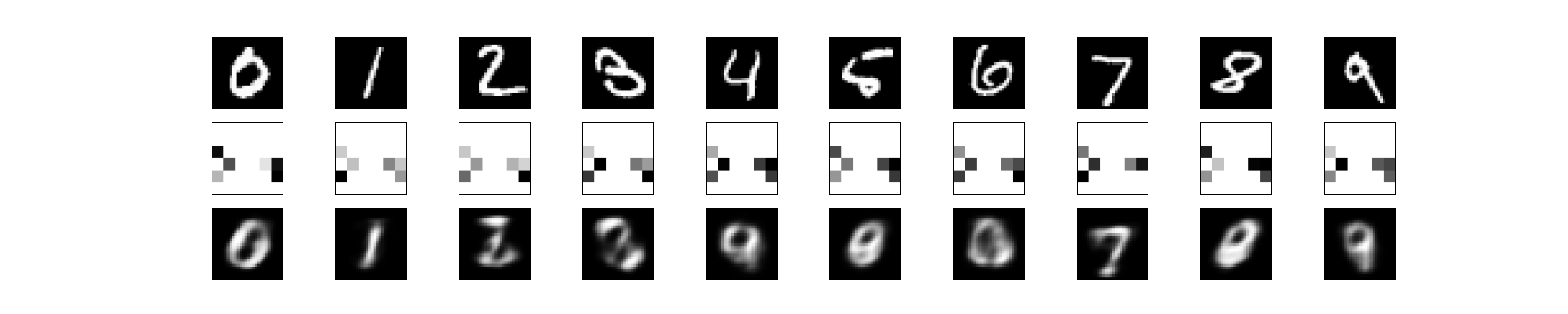}
    \caption{\label{Fig.act.sigm}Sigmoid}
  \end{subfigure}
  \begin{subfigure}{0.48\textwidth}
    \includegraphics[width=\textwidth,trim={14em 2em 14em 19em},clip]{basic-36-tanh-rmsprop-xent.pdf}
    \caption{\label{Fig.act.tanh}Hyperbolic tangent}
  \end{subfigure}

  \vspace{1em}
  
  \begin{subfigure}{0.48\textwidth}
    \includegraphics[width=\textwidth,trim={14em 2em 14em 19em},clip]{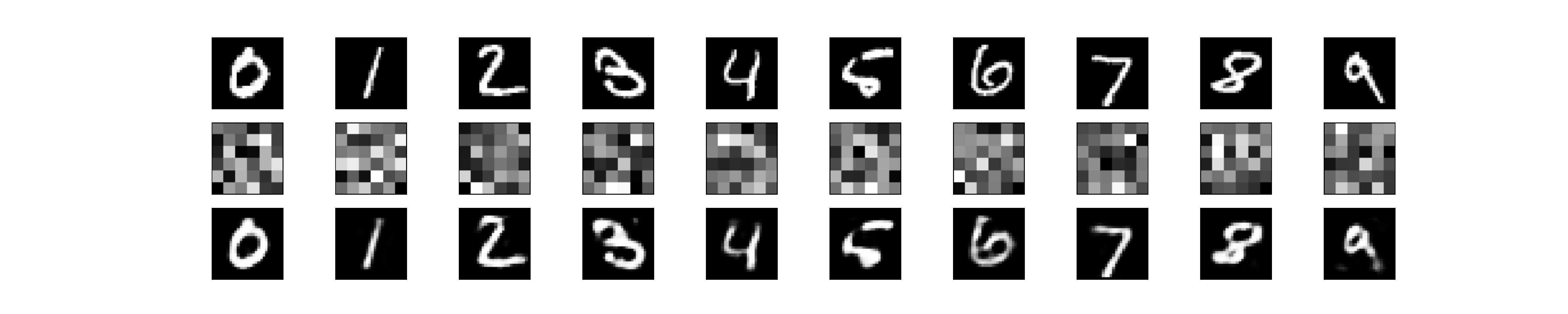}
    \caption{\label{Fig.act.relu}ReLU}
  \end{subfigure}
  \begin{subfigure}{0.48\textwidth}
    \includegraphics[width=\textwidth,trim={14em 2em 14em 19em},clip]{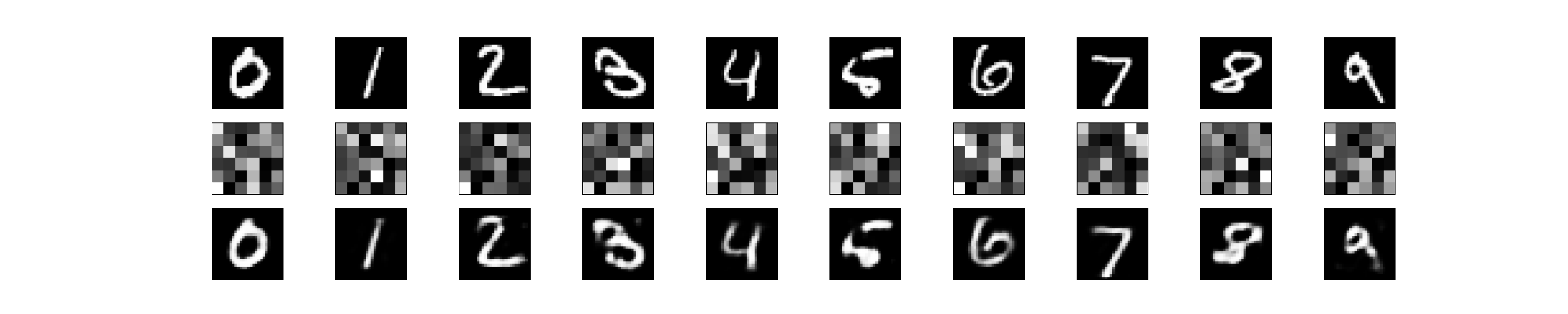}
    \caption{\label{Fig.act.selu}SELU}
  \end{subfigure}
  \caption{\label{Fig.activations}Test samples and reconstructions obtained with different activation functions.}
\end{figure*}

\begin{figure*}[htbp]
  \centering
  \begin{subfigure}{0.48\textwidth}
    \includegraphics[width=\textwidth,trim={14em 0 14em 0},clip]{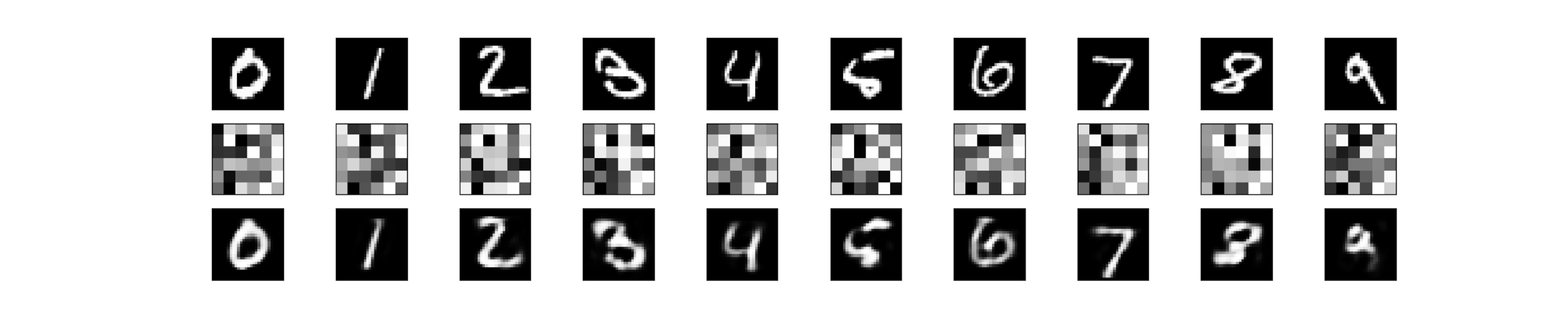}
    \caption{\label{Fig.mnist.basic}Basic AE}
  \end{subfigure}
  \hfill
  \begin{subfigure}{0.48\textwidth}
    \includegraphics[width=\textwidth,trim={14em 0 14em 0},clip]{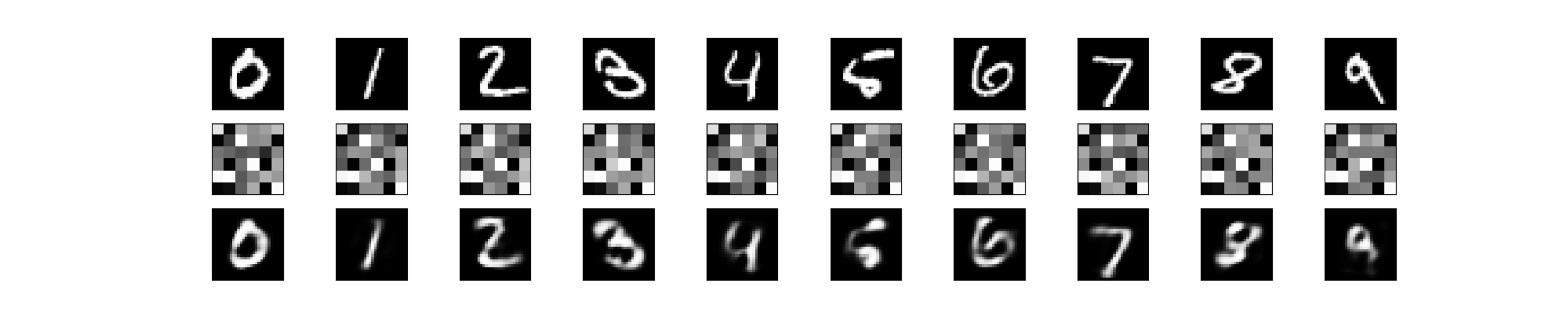}
    \caption{\label{Fig.mnist.wd}Basic AE with weight decay}
  \end{subfigure}

  \begin{subfigure}{0.48\textwidth}
    \includegraphics[width=\textwidth,trim={14em 0 14em 0},clip]{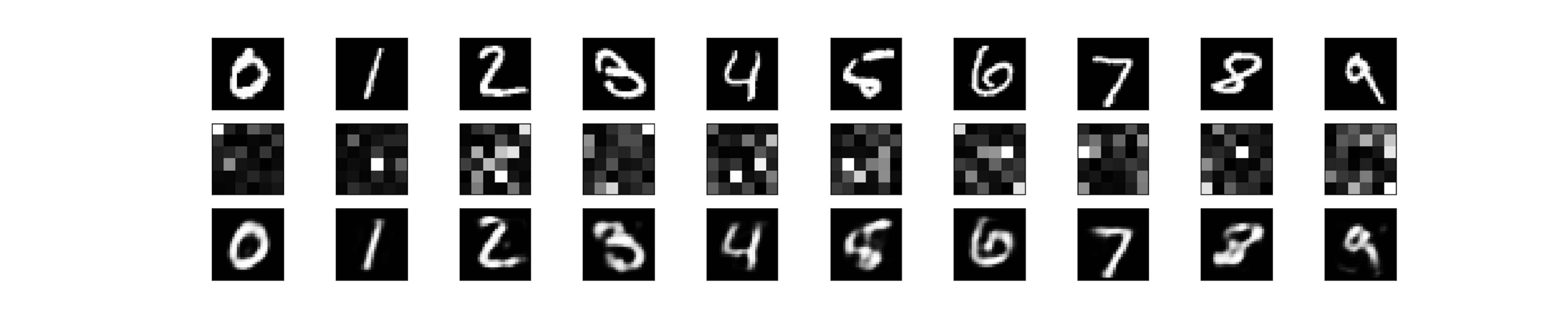}
    \caption{\label{Fig.mnist.sparse}Sparse AE}
  \end{subfigure}
  \hfill
  \begin{subfigure}{0.48\textwidth}
    \includegraphics[width=\textwidth,trim={14em 0 14em 0},clip]{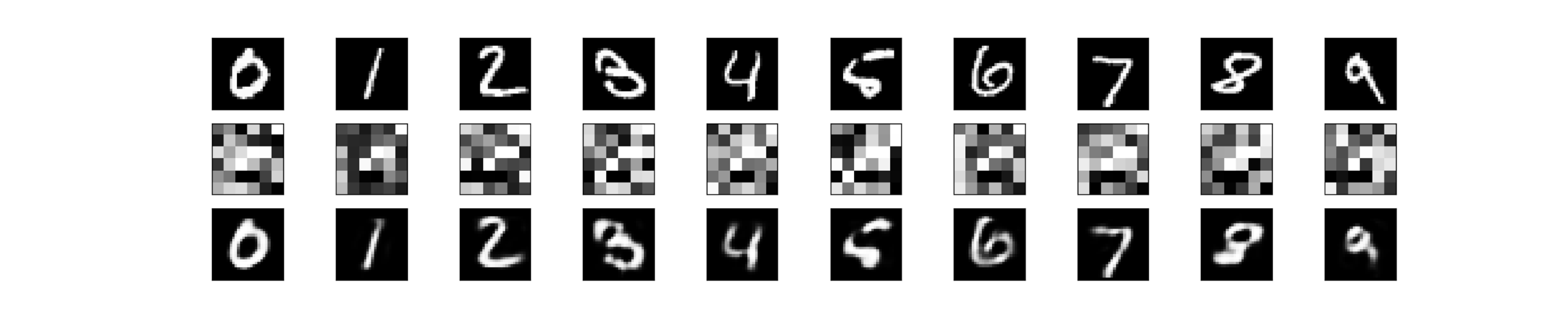}
    \caption{\label{Fig.mnist.contractive}Contractive AE}
  \end{subfigure}
  
  \begin{subfigure}{0.48\textwidth}
    \includegraphics[width=\textwidth,trim={14em 0 14em 0},clip]{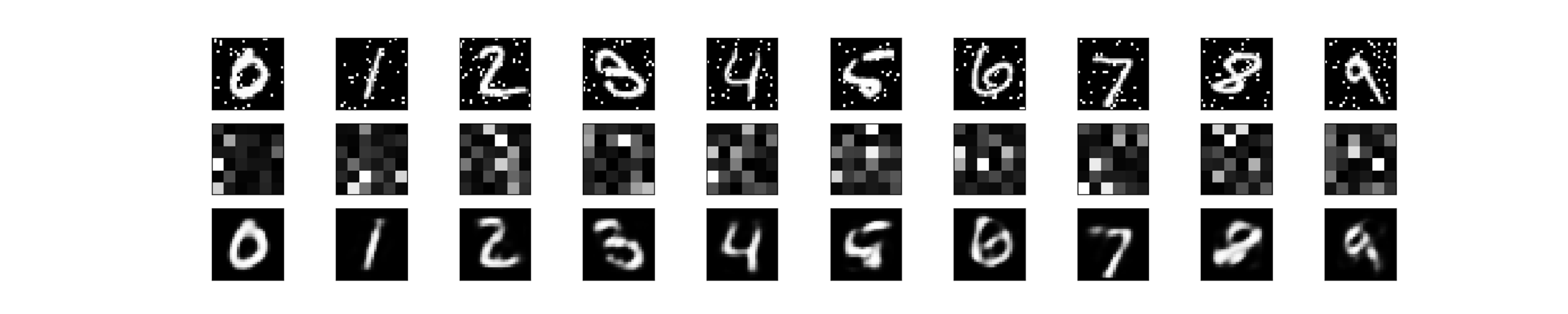}
    \caption{\label{Fig.mnist.denoising}Denoising AE with sparsity regularization}
  \end{subfigure}
  \hfill
  \begin{subfigure}{0.48\textwidth}
    \includegraphics[width=\textwidth,trim={14em 0 14em 0},clip]{robust-wd-36-rmsprop-corr.pdf}
    \caption{\label{Fig.mnist.robust}Robust AE with weight decay}
  \end{subfigure}
  \caption{\label{Fig.mnist}Reconstructing test samples with different AE models. First row of each figure shows test samples, second row shows activations of the encoding layer and third row displays reconstructions. Encoded values range from -1 (black) to 1 (white).}
  
\end{figure*}

In addition, Fig.~\ref{Fig.optimizers} provides the reconstructions generated for some test instances for a basic AE trained with each of those optimizers. As could be intuitively deduced by the convergence, or lack thereof, of the methods, SGD was not capable of finding weights which would recover any digit. AdaGrad, for its part, did improve on SGD but its reconstructions are relatively poor, whereas Adam and RMSProp display superior performance, with little difference between them.

\subsubsection{Comparison of activation functions}

Activation functions play an important role in the way gradients are propagated through the network. In this case, we apply four widely used activations in the encoding layer of a basic AE and compare how they affect its reconstruction ability. Some example results can be seen in Fig.~\ref{Fig.activations}.

Sigmoid and hyperbolic tangent are functions with similar properties, but in spite of this they produce remarkably dissimilar results. Reconstructions are poor when using sigmoidal activation, while $\tanh$ achieves representations much closer to the original inputs.

With respect to ReLU and SELU, the observed results are surprisingly solid and almost indistinguishable. They perform slightly better than $\tanh$ in the sense that reconstructions are noticeably sharper. Their good performance in this case may be due to the nature of the data, which is restricted to the $[0,1]$ interval and does not necessarily show the behavior of these activations in general.

\subsubsection{Comparison of the main AE models}\label{Sec.case.models}

It can be interesting to study the different traits the codifications may acquire when variations on the basic AE are introduced. The reconstructions produced by six different AE models are shown in Fig.~\ref{Fig.mnist}. 

The basic AE (Fig.~\ref{Fig.mnist.basic}) and the one with weight decay (Fig.~\ref{Fig.mnist.wd}) both generate recognizable reconstructions, although slighly blurry. They however do not produce much variability among different digits in the encoding layer, which means they are not making full use of its 36 dimensions. The weight decay corresponds to Eq.~\ref{Eq.wd} with $\lambda$ set to $0.01$.

The sparse AE has been trained according to Eq.~\ref{Eq.KLdivergence} with an expected activation value of $-0.7$. Its reconstructions are not much different from those of the previous ones, but in this case the encoding layer has much lower activations in average, as can be appreciated by the darker representations in  Fig.~\ref{Fig.mnist.sparse}. Most of the information is therefore tightly condensed in a few latent variables.

The contractive AE achieves other interesting properties in its encoding: it has attempted to model the data as a lower dimensional manifold, where digits that seem more similar will be placed closer than those which are very unalike. As a consequence, the 0 and the 1 shown in Fig.~\ref{Fig.mnist.contractive} have very differing codifications, whereas the 3 and the 8 have relatively similar ones. Intuitively, one would need to travel larger distances along the learned manifold to go from a 0 to a 1, than from a 3 to an 8.

The denoising AE is able to eliminate noise from test instances, at the expense of losing some sharpness in the reconstruction, as can be seen in Fig.~\ref{Fig.mnist.denoising}. Finally, the robust AE (Fig.~\ref{Fig.mnist.robust}) achieves noticeably higher clarity in the reconstruction and more variance in the encoding than the standard AEs.

\section{Conclusions}\label{Conclusions}
As Pedro Domingos states in his famous tutorial \cite{Domingos2012AFU}, and as can be seen from the large number of publications on the subject, feature engineering is the key to obtain good machine learning models, able to generalize and provide decent performance. This process consists in choosing the most relevant subset of features or combining some of them to create new ones. Automated fusion of features, specially when performed by nonlinear techniques, has demonstrated to be very effective. Neural network-based autoencoders are among the approaches to conduct this kind of task.

This paper started offering the reader with a general view of which an AE is, as well as its essential foundations. After introducing the usual AE network structures, a new AE taxonomy, based on the properties of the inferred model, has been proposed. Those AE models mainly used in feature fusion have been explained in detail, highlighting their most salient characteristics and comparing them with more classical feature fusion techniques. The use of disparate activation functions and training methods for AEs has been also thoroughly illustrated. 

In addition to AEs for feature fusion, many other AE models and applications have been listed. The number of new proposals in this field is always growing, so it is easy to find dozens of AE variants, most of them based on the fundamental models described above.

This review is complemented by a final section proposing guidelines for selecting the most appropriate AE model based on different criteria, such as the type of units, loss function, activation function, etc., as well as mentioning available software to put this knowledge into practice. Empirical results on the well known MNIST dataset obtained from several AE configurations, combining disparate activation functions, optimizers and models, have been compared. The aim is to offer the reader help when facing this type of decision.

\textbf{Acknowledgments}:  This work is supported by the Spanish National Research Projects TIN2015-68454-R and  TIN2014-57251-P, and Project BigDaP-TOOLS - Ayudas Fundaci\'on BBVA a Equipos de Investigaci\'on Cient\'ifica 2016. 


\bibliographystyle{elsarticle-num}
\bibliography{references}

\clearpage

\appendix
\section{Description of used datasets}

\subsection{Breast Cancer Diagnosis (Wisconsin)}

The well known dataset of diagnosis of breast cancer in Wisconsin (WDBC) \cite{WDBC} is briefly used in Section~\ref{Sec.Visualization} to provide a 2-dimensional visualization example.

This dataset consists of 569 instances corresponding to patients, each of which present 30 numeric input features and one of two classes that identify the type of tumor: benign or malignant. The dataset is slightly imbalanced, exhibiting a 37.3\% of instances associated to the malignant class, while the remaining 62.7\% correspond to benign tumors. The data have been normalized for the training process of the basic AE that generated the example.

Originally, features were extracted from a digitized image of a fine-needle aspiration sample of a breast mass, and described ten different traits of each cell nucleus. The mean, standard error and largest value of these features are computed, resulting in the 30 input attributes for each patient, gathered in the published dataset.

WDBC is usually relied on as an example dataset and most classifiers generally obtain high accuracy: the authors of the original proposal already achieved 97\% of classification accuracy in cross-validation. However, it presents some issues when applying AEs: its small imbalance may cause instances classified as benign to contribute more to the loss function, inducing some bias in the resulting network, which may reconstruct these more accurately than the rest. Furthermore, it is composed of relatively few instances, which may not be sufficient for some deep learning techniques to be able to generalize.

\subsection{MNIST}

MNIST \cite{MNIST} is a widely used dataset within deep learning research. It is regularly chosen as a benchmark for new techniques and neural architectures. It has been the base of our case study in Section~\ref{Sec.CaseStudy}.

The dataset consists of 60~000 instances, divided into a 50~000-instance set for training and the remaining 10~000 for test. each corresponding to a 28x28-sized image of a handwritten digit, from 0 to 9. The values of this 28x28 matrix or 784-variable input represent the gray level of each pixel, and therefore range from 0 to 255, but they have been rescaled to the $[0,1]$ interval in our examples.

This dataset is actually a modified subset of a previous work from NIST\footnote{Available at \url{http://doi.org/10.18434/T4H01C}.} for character recognition. The original images used only black or white pixels, whereas in MNIST they have been anti-aliased.

MNIST has been used as benchmark for a large variety of deep learning proposals, since it is reasonably easy to extract higher-level features out of simple grayscale images, and it provides a high enough amount of training data. State-of-the-art work\footnote{A collection of methods applied to MNIST and their results is available at \url{http://yann.lecun.com/exdb/mnist/}.} achieves an error rate of around 0.2\% \cite{Alvear_Sandoval_2018,MulticolumnDNN}.

\end{document}